\documentclass[10pt,twocolumn,letterpaper]{article}

\usepackage[final]{iccv}      
\usepackage{bm,algorithm,algorithmic}

\definecolor{iccvblue}{rgb}{0.21,0.49,0.74}
\usepackage[pagebackref,breaklinks,colorlinks,allcolors=iccvblue]{hyperref}

\def\paperID{3187} 
\def\confName{ICCV}
\def\confYear{2025}

\def\method{ULTHO }
\def\nsmethod{ULTHO}

\title{ULTHO: Ultra-Lightweight yet Efficient Hyperparameter Optimization in\\Deep Reinforcement Learning}

\author{Mingqi Yuan\textsuperscript{1}, Bo Li\textsuperscript{1}, Xin Jin\textsuperscript{2,3}\thanks{Corresponding author}, Wenjun Zeng\textsuperscript{2,3}\\
\textsuperscript{1}Department of Computing, The Hong Kong Polytechnic University, Hong Kong SAR, China\\
\textsuperscript{2}Ningbo Institute of Digital Twin, Eastern Institute of Technology, Ningbo, China\\
\textsuperscript{3}Ningbo Key Laboratory of Spatial Intelligence and Digital Derivative, Ningbo, China
}

\begin{document}
\maketitle

\begin{abstract}
Hyperparameter optimization (HPO) is a billion-dollar problem in machine learning, which significantly impacts the training efficiency and model performance. However, achieving efficient and robust HPO in deep reinforcement learning (RL) is consistently challenging due to its high non-stationarity and computational cost. To tackle this problem, existing approaches attempt to adapt common HPO techniques (e.g., population-based training or Bayesian optimization) to the RL scenario. However, they remain sample-inefficient and computationally expensive, which cannot facilitate a wide range of applications. In this paper, we propose ULTHO, an ultra-lightweight yet powerful framework for fast HPO in deep RL within single runs. Specifically, we formulate the HPO process as a multi-armed bandit with clustered arms (MABC) and link it directly to long-term return optimization. ULTHO also provides a quantified and statistical perspective to filter the HPs efficiently. We test ULTHO on benchmarks including ALE, Procgen, MiniGrid, and PyBullet. Extensive experiments demonstrate that the ULTHO can achieve superior performance with a simple architecture, contributing to the development of advanced and automated RL systems. Our code is available at the GitHub repository\footnote{\url{https://github.com/yuanmingqi/ULTHO}}.

\end{abstract}

\section{Introduction}
Deep reinforcement learning (RL) has propelled significant advancements across various fields, such as game playing \cite{mnih2015human,silver2016mastering,vinyals2019grandmaster}, chip design \cite{goldie2024addendum}, algorithm innovation \cite{fawzi2022discovering,mankowitz2023faster}, and large language model (LLM) \cite{ouyang2022training,bai2022training,bi2024deepseek,liu2024deepseek}. However, training deep RL agents involves a number of design decisions and hyperparameter configurations, in which minor variations can result in substantial effects on learning efficiency and final performance \cite{eimer2023hyperparameters}. While many efforts have been devoted to improving the design choices \cite{henderson2018deep,engstrom2020implementation,hsu2020revisiting,huang202237} and implementation details \cite{stable-baselines3,andrychowicz2021what,huang2022cleanrl,yuan2025rllte}, RL-specific hyperparameter optimization (HPO) does not receive sufficient attention in the community. This gap is particularly notable given the high sensitivity of deep RL algorithms to HPs, which restricts their broader applications and the development of advanced and automated RL systems \cite{franke2021sampleefficient}.
\begin{figure}
\begin{center}
\includegraphics[width=\linewidth]{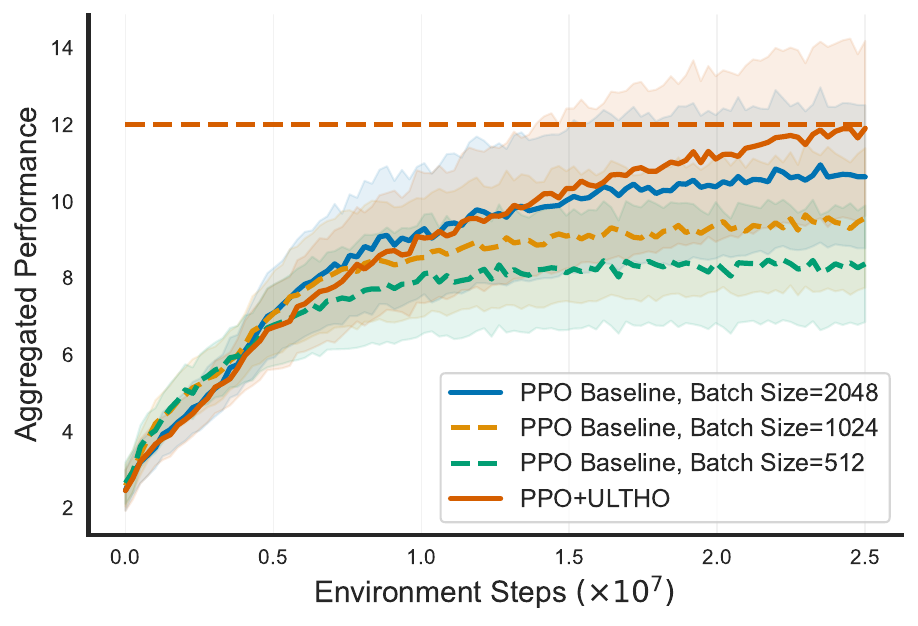}
\end{center}
\vspace{-20pt}
\caption{Aggregated performance comparison on the Procgen benchmark with procedurally-generated environments. The choice of batch size significantly impacts the agent's performance. \method can effectively perform HPO across different tasks and learning stages, thereby improving the overall performance.}
\label{fig:preface}
\vspace{-15pt}
\end{figure}

HPO is a fundamental problem in machine learning, with traditional methods like grid search \cite{yang2020hyperparameter} and random search \cite{bergstra2012random} being widely used to automate the process. More advanced techniques, such as Bayesian optimization \cite{snoek2012practical} and multi-fidelity approaches like Hyperband \cite{li2018hyperband}, have been developed to improve efficiency by leveraging surrogate models and early stopping mechanisms. However, these methods are primarily designed for static optimization problems, where the optimal HPs remain fixed throughout the training process. In contrast, deep RL operates in non-stationary environments where the optimal HPs may vary across different training stages, especially in dynamic environments such as Procgen \cite{cobbe2020leveraging} and Craftax \cite{matthews2024craftax}. This non-stationarity, along with the high computational cost of deep RL training, makes traditional HPO methods less effective for these applications.

To tackle the HPO challenge in deep RL, several approaches have been proposed to dynamically adjust HPs during training. For instance, \cite{jaderberg2017population} introduced population-based training (PBT) that evolves a population of agents to explore and exploit different HP schedules, enabling adaptation to non-stationary settings. Notably, PBT is utilized to solve the complex 3D multiplayer game, Quake III Arena, achieving human-level performance by optimizing the agent policies along with internal reward signals in a hierarchical manner \cite{jaderberg2019human}. \cite{franke2021sampleefficient} further extended PBT and proposed SEARL, which optimizes the HPs and also the neural network architecture while simultaneously training the agent, significantly improving sample efficiency by sharing experience across the population. However, PBT-like methods remain sample-inefficient and computationally expensive, limiting their practicality to a wider range of tasks. To address these issues, single-run HPO methods such as HOOF \cite{paul2019fast} have been developed. HOOF adapts HPs using off-policy importance sampling, optimizing a one-step improvement objective with sampled trajectories. As a gradient-free algorithm, HOOF can significantly reduce overhead and promote sample efficiency. However, HOOF greedily focuses on HP configurations that maximize the value of the updated policy without introducing an exploration mechanism. Moreover, its reliance on importance sampling also makes it prone to high variance when policy distributions shift, hurting its robustness in highly dynamic environments.



Inspired by the discussions above, we propose \textbf{\method}: \textbf{U}ltra-\textbf{L}ightweigh\textbf{T} \textbf{H}yperparameter \textbf{O}ptimization, a general, simple, yet powerful framework for achieving efficient and robust HPO in deep RL within single runs. Unlike previous methods that focus on short-term improvements or require complex learning processes, ULTHO performs HPO from the perspective of the hierarchical bandits, ensuring long-term performance gains with minimal computational overhead. Our main contributions are summarized as follows:
\begin{itemize}
    \item  We formulate the HPO process as a multi-armed bandit with clustered arms (MABC), which optimizes HPs adaptively across different tasks and learning stages in a hierarchical fashion and effectively reduces the sample complexity. \method selects the appropriate HPs based on the estimated task return, ensuring the maximization of long-term returns while balancing exploration;
    \item  Our framework currently provides two algorithms, {\em i.e.}, a normal version and an extended version for continual optimization, solving HPO within single runs and providing a quantified and statistical perspective to analyze the potential of distinct HPs. In particular, \method has a simple architecture and requires no additional learning processes, which can be compatible with a broad range of RL algorithms;
    \item  Finally, we evaluate \method on ALE (arcade game environments), Procgen (sixteen procedurally-generated environments), MiniGrid (environments with sparse rewards), and PyBullet (robotics environments with continuous action space). Extensive experiments demonstrate that \method can achieve superior performance with remarkable computational efficiency.

\end{itemize}

\section{Related Work}
\subsection{HPO in Machine Learning}
HPO is essential in machine learning as it significantly impacts model performance, convergence speed, and generalization ability. Prominent techniques, including random and grid search \cite{bergstra2012random}, Bayesian optimization (BO) \cite{snoek2012practical,wu2019hyperparameter,victoria2021automatic}, multi-fidelity search strategies \cite{li2018hyperband,falkner2018bohb,jiang2024efficient}, grdient-based methods \cite{maclaurin2015gradient,xu2018meta,bohdal2021evograd,micaelli2021gradient}, population-based training \cite{jaderberg2017population,parker2020provably,wan2022bayesian}, and RL-based methods \cite{dong2019dynamical,jomaa2019hyp,iranfar2021multiagent}. For instance, \cite{falkner2018bohb} proposes BOHB that combines BO and Hyperband to balance the exploration-exploitation trade-off while dynamically allocating resources, significantly improving efficiency and robustness over a single method. \cite{jomaa2019hyp} further extends PBT and proposed population-based bandits (PB2) by incorporating a MAB strategy to adaptively explore HP schedules, improving sample efficiency over standard PBT. \cite{jomaa2019hyp} formulates HPO as a sequential decision problem and solves it with RL, which enables adaptive HP tuning and reduces reliance on hand-crafted acquisition functions. However, the methods above are prone to be computationally expensive and less effective for RL scenarios.

In this paper, we solve the HPO problem from the perspective of hierarchical bandits, achieving efficient scheduling within single runs and significantly reducing the computational overhead.




\begin{figure*}[t!]
\begin{center}
\includegraphics[width=0.95\linewidth]{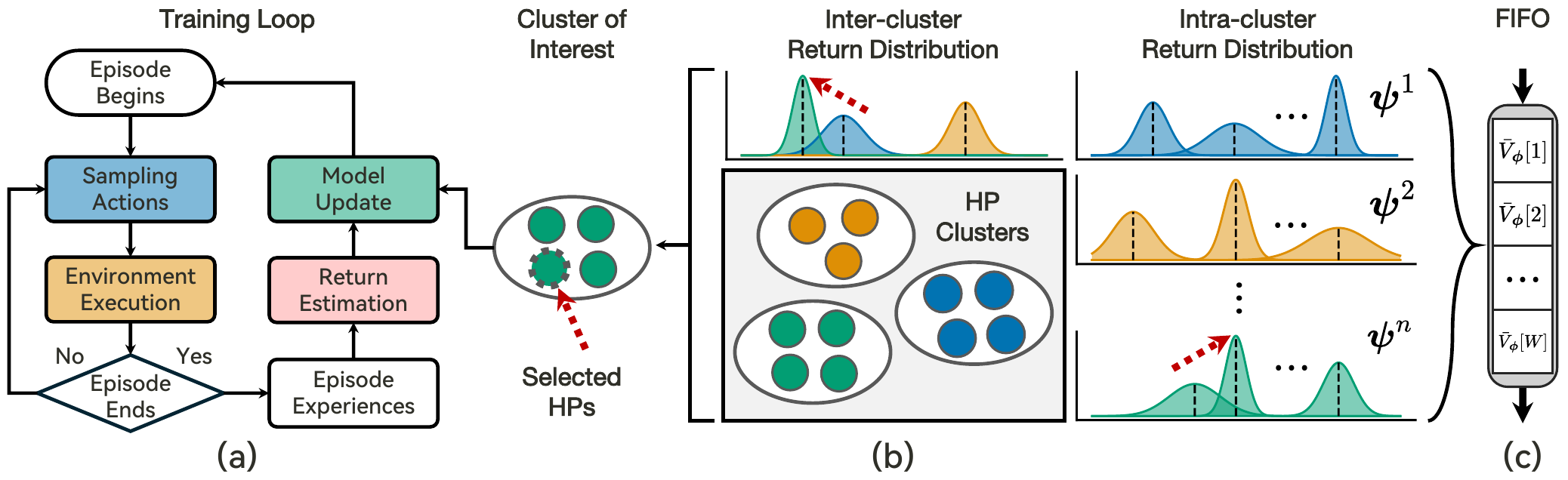}
\end{center}
\vspace{-15pt}
\caption{Overview of \method framework. (a) The key phases of RL algorithms. \method serves as a plug-and-play module that feeds the optimized HPs to the RL algorithm. (b) \method maintains inter-cluster and intra-cluster return distributions to perform HPO in a hierarchical manner. (c) A sliding window is used to store the estimated task returns for updating the return distributions.}
\label{fig:overview}
\vspace{-15pt}
\end{figure*}

\subsection{HPO in Reinforcement Learning}
Extensive research \cite{henderson2018deep,engstrom2020implementation,zhang2021importance,huang202237,eimer2023hyperparameters,tejer2024importance} has shown the importance of HPO in deep RL. However, directly applying HPO methods from general machine learning to deep RL is consistently inefficient due to its non-stationarity and high computational cost. One promising approach to address this is meta-gradients \cite{xu2018meta}, which dynamically adjust HPs during training. For instance, \cite{zahavy2020self} uses meta-gradient descent to self-tune all the differentiable HPs of an actor-critic loss function and discover auxiliary tasks, while improving off-policy learning using a novel leaky V-trace operator. \cite{flennerhag2022bootstrapped} further enhances this by bootstrapping the meta-learning process, allowing agents to meta-learn more efficiently across tasks and providing robust performance in dynamic environments. However, these methods require access to the algorithms' gradients and result in more computational overhead, limiting their use in common RL settings.

In this paper, we optimize HPs based solely on the estimated task returns, avoiding complex learning processes and not requiring access to internal data such as gradients, thereby facilitating a wider range of RL algorithms.



\subsection{MAB Algorithms for Deep RL}
MAB problems are closely related to RL, as both involve decision-making under uncertainty \cite{auer2002finite}. While RL focuses on sequential decisions to maximize cumulative rewards, MAB methods optimize immediate actions, making them effective for addressing subproblems within RL frameworks. For example, \cite{raileanu2020automatic} proposed UCB-DrAC, which employs a bandit algorithm to select optimal data augmentations, significantly improving generalization in procedurally-generated environments. Similarly, AIRS \cite{yuan2023automatic} formulates intrinsic reward selection as a bandit problem, dynamically adapting rewards to enhance exploration at different learning stages. Finally, \cite{yuan2025adaptive} applies bandit algorithms to adaptively control the data exploitation, enhancing the data efficiency and generalization while reducing the overall computational overhead.

In this paper, our framework unifies the methods above by extending the MAB problem with clustered arms, enabling more efficient HPO via hierarchical exploration.

\section{Background}
\subsection{Reinforcement Learning}
We study the RL problem considering a Markov decision process (MDP) \cite{bellman1957markovian, kaelbling1998planning} defined by a tuple $\mathcal{M}=(\mathcal{S},\mathcal{A},r,P,d_{0},\gamma)$, where $\mathcal{S}$ is the state space, $\mathcal{A}$ is the action space, and $r:\mathcal{S}\times\mathcal{A}\rightarrow\mathbb{R}$ is the extrinsic reward function, $P:\mathcal{S}\times\mathcal{A}\rightarrow\Delta(\mathcal{S})$ is the transition function that defines a probability distribution over $\mathcal{S}$, $d_{0}\in\Delta(\mathcal{S})$ is the distribution of the initial observation $\bm{s}_{0}$, and $\gamma\in[0, 1]$ is a discount factor. The goal of RL is to learn a policy $\pi_{\bm\theta}(\bm{a}|\bm{s})$ to maximize the expected discounted return:
\begin{equation}
	J_{\pi}(\bm{\theta})=\mathbb{E}_{\pi}\left[\sum_{t=0}^{\infty}\gamma^{t}r(\bm{s}_t,\bm{a}_t)\right].
\end{equation}
    

\subsection{HPO in RL}
HPO in RL aims to find the optimal set of HPs that maximizes the expected return of the agent's policy \cite{franke2021sampleefficient,eimer2023hyperparameters}. In this context, HPs refer to parameters that are not directly learned by the RL algorithm but influence the learning process, such as learning rates, exploration parameters, and network architecture choices. The objective of HPO in RL can be formulated as solving the following black-box optimization problem:
\begin{equation} 
\bm{\psi}^{*}=\underset{\bm{\psi}\in\bm{\Psi}}{\rm argmax}\:J_{\pi}(\bm{\theta}
,\bm{\psi},\boldsymbol{\mathsf{A}}),
\end{equation}
where $\bm{\psi}^{*}$ is the optimal configuration, $\bm{\Psi}$ is the HP space, and $\boldsymbol{\mathsf{A}}
$ is an RL algorithm.

\section{The \method Framework}
In this section, we introduce ULTHO, an ultra-lightweight, unified, yet powerful framework designed for efficient and robust HPO in RL, whose overview is presented in Figure~\ref{fig:overview}. Our key insights are as follows: (i) Similar to \cite{paul2019fast} and \cite{zahavy2020self}, \method also dynamically tunes HPs within a single training run, enabling adaptation to different learning stages and significantly reducing the trial-and-error cost; (ii) \method adopts a hierarchical approach \cite{jaderberg2019human,barsce2020hierarchical} to organize and select the HP candidates. Specifically, we perform a two-tier HPO to traverse the search space more efficiently; (iii) Unlike prior methods \cite{jaderberg2017population,paul2019fast,zahavy2020self} often prioritizes short-term improvements, \method focuses on optimizing long-term returns, providing a systematic approach that ensures more sustainable performance gains; (iv) To facilitate a wide range of RL algorithms, \method requires no additional learning processes or access to internal data, such as gradients, making it universally applicable without requiring complex modifications to the underlying algorithm. 

\subsection{MABC for HPO}
To simplify the notations, we reuse $\bm{\Psi}$ to denote the set of HP clusters: 
\begin{equation}
    \bm{\Psi}=\{\bm{\psi}^{1},\bm{\psi}^{2},\dots,\bm{\psi}^{n}\},
\end{equation}
where 
\begin{equation}
    \bm{\psi}^{i}=\{\psi_{1}^{i},\psi_{2}^{i},\dots,\psi_{m}^{i}\}
\end{equation}
is an individual cluster of HPs.

Then, the HPO at different learning stages can be formulated as a multi-armed bandit with clustered arms (MABC) \cite{carlsson2021thompson}, and the optimization objective is to maximize the long-term return evaluated by the task reward function.

\begin{algorithm}[t!]
	\caption{The \method with UCB}
	\label{algo:main}
	\begin{algorithmic}[1]
    \STATE Initialize the policy network $\pi_{\bm \theta}$ and value network $V_{\bm \phi}$;
    \STATE Initialize a set $\bm{\Psi}$ of HPs, an exploration coefficient $c$, a window length $W$ for estimating the utility functions;
    \STATE $\forall \bm{\psi}\in\bm{\Psi},\forall \psi\in\bm{\psi}$, let
    \begin{align*}
            N(\bm{\psi})=1, U(\bm{\psi})=0, R(\bm{\psi})={\rm FIFO}(W)\\N(\psi)=1, U(\psi)=0, R(\psi)={\rm FIFO}(W)
    \end{align*}
    \vspace{-12pt}
    \FOR{each episode $e$}{
        \STATE Sample rollouts using the policy network $\pi_{\bm \theta}$;
        \STATE Perform the generalized advantage estimation (GAE) to get the estimated task returns;
        \STATE Select a cluster $\bm{\psi}_e$ using Eq.~(\ref{eq:ucb_inter});
        \STATE Select a HP $\psi_e$ from the $\bm{\psi}_e$ using Eq.~(\ref{eq:ucb_intra});
        \STATE Update policy network and value network;
        \STATE Compute $\bar{V}_{\bm\phi}$ obtained by the new policy;
        \STATE Add $\bar{V}_{\bm\phi}$ to the queue $R(\bm{\psi}_{e})$ and $R(\psi_{e})$;
        \STATE Update $U(\bm{\psi}_e)$ and $U(\psi_e)$ using Eq.~(\ref{eq:aer});
        \STATE $N(\bm{\psi}_{e})\leftarrow N(\bm{\psi}_{e})+1,N(\psi_{e})\leftarrow N(\psi_{e})+1$.
        }
	\ENDFOR
	\end{algorithmic}
\end{algorithm}

We solve the defined MABC problem by hierarchically applying the upper confidence bound (UCB) \cite{auer2002using} algorithm, which is an effective and widely-used method to balance exploration-exploitation in bandit problems. Specifically, at the $i$-th episode, we first select a cluster from $\bm{\Psi}$ by the following policy:
\begin{equation}\label{eq:ucb_inter}
    \underbrace{\bm{\psi}_{i}=\underset{ \bm{\psi}\in\bm{\Psi}}{\rm argmax}\left[U_{i}( \bm{\psi})+c\sqrt{\frac{\log i}{N_{i}(\bm{\psi})}}\right]}_{\text{Inter-cluster Scheduling Policy}},
\end{equation}
where $N_{i}(\bm{\psi})$ is the number of times that $\bm{\psi}$ has been chosen before the $i$-the episode, and $c$ is the exploration coefficient. 

Equipped with the selected cluster, we then select a specific HP value following a similar policy:
\begin{equation}\label{eq:ucb_intra}
    \underbrace{\psi_{i}=\underset{\psi\in\bm{\psi}_{i}}{\rm argmax}\left[U_{i}(\psi)+c\sqrt{\frac{\log i}{N_{i}(\psi)}}\right]}_{\text{Intra-cluster Scheduling Policy}}.
\end{equation}

The selected HP value will be used for policy updates. Then the utility functions of the cluster and the corresponding value are updated using a sliding window average of the past mean episode returns obtained by the agent:
\begin{equation}\label{eq:aer}
    \underbrace{U_{i}(\bm{\psi})=\frac{1}{W}\sum_{j=1}^{W}\bar{V}_{\bm \phi}[j],\bar{V}_{\bm \phi}=\frac{1}{T}\sum_{t=1}^{T}V_{\bm \phi}(\bm{s}_{t})}_{\text{Utility Evaluation}},
\end{equation}
where $V_{\bm\phi}(\bm{s}_{t})$ is the estimated return predicted by the value network, $T$ is the episode length, and $W$ is the window length.


\begin{algorithm}[t!]
	\caption{The Relay-\method}
	\label{algo:relay}
	\begin{algorithmic}[1]
    \STATE Execute the Algorithm~\ref{algo:main};
    \STATE Get the $\bm{\psi^{\rm COI}}$ and $\bm{\psi^{\rm NOI}}$ using Eq.~(\ref{eq:coi});
    \STATE Execute Algorithm~\ref{algo:main} solely with $\bm{\psi^{\rm COI}}$ or $\bm{\psi^{\rm NOI}}$;
    \STATE Output the best-performing cluster.
	\end{algorithmic}
\end{algorithm}

\subsection{Relay-\method}
Algorithm~\ref{algo:main} summarizes the workflow of \method with UCB. It is evident that the \method changes only one HP at a time, which helps stabilize the learning process due to the high sensitivity of RL to HPs. However, optimizing multiple HPs simultaneously can lead to unstable learning behavior, slower convergence, and potentially sub-optimal policies in certain environments. Moreover, it is unclear whether all HPs consistently affect performance across different environments and learning stages. With this in mind, we propose an extended algorithm entitled Relay-\nsmethod, as depicted in Algorithm~\ref{algo:relay}.

The core idea behind Relay-\method is to identify two HP clusters: the cluster of interest (COI) and the neglected cluster of interest (NOI). COI is the cluster that is selected most frequently, indicating it has the greatest impact on the agent’s performance. In contrast, NOI is the cluster with the least number of selections. After executing Algorithm~\ref{algo:main} with an initial set of clusters $\bm{\Psi}$, we compute the total count $N_{\rm end}(\bm{\psi})$ for each cluster. Then we define the COI and NOI as
\begin{equation}\label{eq:coi}
\begin{aligned}
    \bm{\psi}^{\rm COI}&=\underset{\bm{\psi}\in\bm{\Psi}}{\rm argmax}\:N_{\rm end}
    (\bm{\psi}),\\
    \bm{\psi}^{\rm NOI}&=\underset{\bm{\psi}\in\bm{\Psi}}{\rm argmin}\:N_{\rm end}
    (\bm{\psi}).
\end{aligned}
\end{equation}

Once the COI and NOI are identified, the optimization process is focused on these clusters, allowing the algorithm to refine the most influential cluster and further explore the neglected cluster. This approach can improve the completeness of exploration, thereby squeezing out additional performance gains. However, we also need to highlight that this algorithm may result in sub-optimal performance due to its restricted landscape. 

\section{Experiments}

In this section, we design the experiments to investigate the following questions:
\begin{itemize}
    \item \textbf{Q1}: Can \method improve performance as compared to using fixed HP values? (See Figure~\ref{fig:ale_ppo}, \ref{fig:pg_ppo_8}, \ref{fig:pb_ppo}, \ref{fig:mg_ppo_agg}, \ref{fig:pg_ppo_16}, \ref{fig:ale_100m}, and Table~\ref{tb:pg_test})
    \vspace{-5pt}
    \item \textbf{Q2}: Can the relay optimization further enhance the \method algorithm? (See Figure~\ref{fig:ale_ppo}, \ref{fig:pg_ppo_8}, \ref{fig:pb_ppo}, \ref{fig:mg_ppo_agg}, \ref{fig:pg_ppo_16}, \ref{fig:ale_100m}, and Table~\ref{tb:pg_test})
    \vspace{-5pt}
    \item \textbf{Q3}: What are the detailed decision processes of the \method algorithm? (See Figure~\ref{fig:pg_ppo_decision_agg}, \ref{fig:pg_ppo_inter_ct_agg}, \ref{fig:pg_ppo_inter_decisions_16}, and \ref{fig:pg_ppo_inter_ct_16})
    \vspace{-5pt}
    \item \textbf{Q4}: How does \method behave in sparse-rewards environments and continuous control tasks? (See Figure~\ref{fig:pb_ppo} and \ref{fig:mg_ppo_agg})
    \vspace{-5pt}
    \item \textbf{Q5}: How robust is the \method algorithm? (See Figure~\ref{fig:pg_grid})
\end{itemize}

\subsection{Setup}
\subsubsection{Benchmark Selection}
We first evaluate the \method using the arcade learning environment (ALE) benchmark \cite{bellemare2013arcade}, a collection of arcade game environments that requires the agent to learn motor control directly from images. Specifically, we focus on a subset of ALE known as ALE-5, which typically produces median score estimates for all games that are within 10\% of their true values \cite{aitchison2023atari}. For all environments, we stack 4 consecutive frames to form an input state with the data shape of $(84,84,4)$. Additionally, we introduce the Procgen benchmark with 16 procedurally-generated environments. Procegn is similar to the ALE benchmark yet involves much higher dynamicity and presents a more difficult challenge for HPO. All the environments use a discrete fifteen-dimensional action space and generate $(64,64,3)$ RGB observations, and we use the \textit{easy} mode and train the agents on 200 levels before testing them on the full distribution of levels. Furthermore, we introduce the MiniGrid \cite{MinigridMiniworld23} and PyBullet \cite{coumans2016pybullet} to test \method in sparse-rewards environments and continuous control tasks.

\begin{figure}[t!]
\begin{center}
\includegraphics[width=\linewidth]{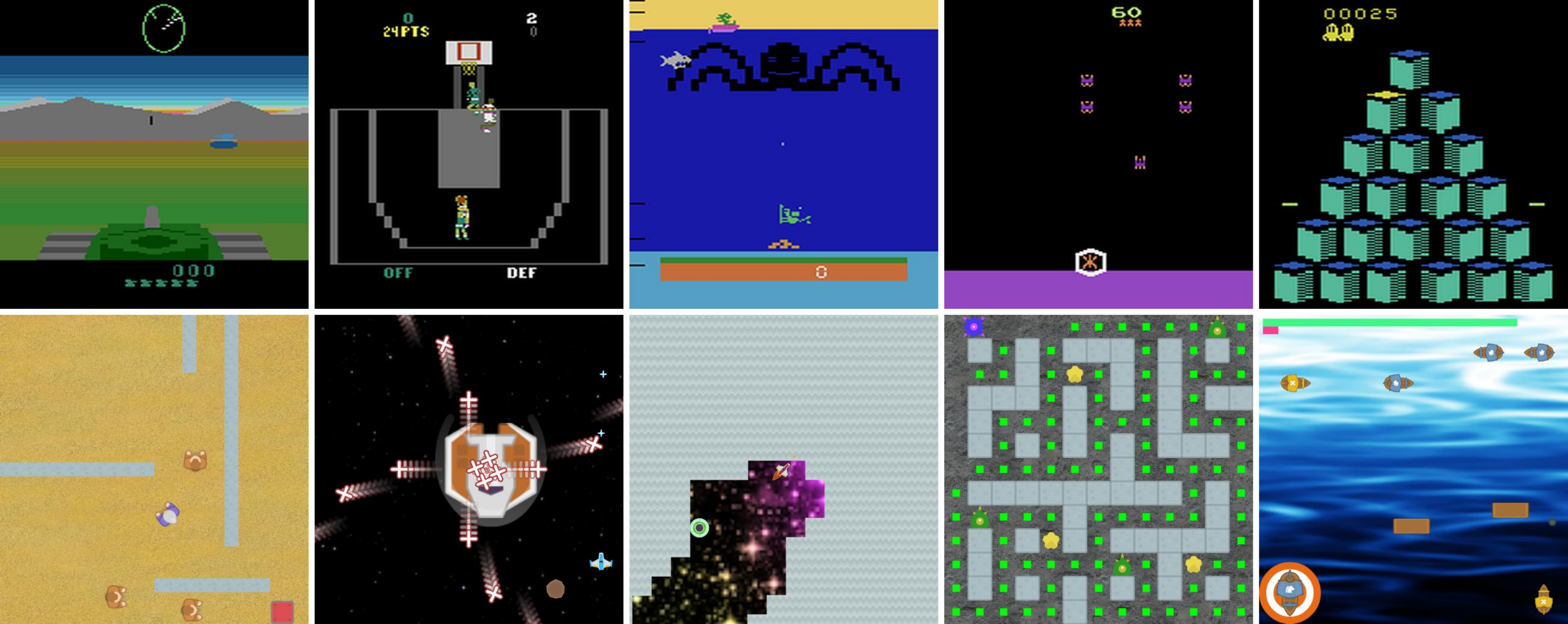}
\end{center}
\vspace{-15pt}
\caption{Screenshots of the ALE (top) and Procgen (below) benchmarks.}
\label{fig:screenshots}
\vspace{-10pt}
\end{figure}

\subsubsection{Algorithmic Baselines}
For the RL algorithm, we select the proximal policy optimization (PPO) \cite{schulman2017proximal} as the baseline, which is a representative algorithm that produces considerable performance on most existing RL benchmarks. For the HPO algorithms, we select random search (RS) \cite{bergstra2012random}, PBT \cite{jaderberg2017population}, PB2 \cite{parker2020provably}, the Bayesian optimization tool SMAC \cite{lindauer2022smac3}, the combination of SMAC and the Hyperband scheduler (SAMC+HB) \cite{li2018hyperband}, and TMIHF \cite{parker2021tuning}. The details of the selected algorithmic baselines can be found in Appendix~\ref{appendix:baseline}.


\begin{figure*}[t!]
  \centering
   \includegraphics[width=\linewidth]{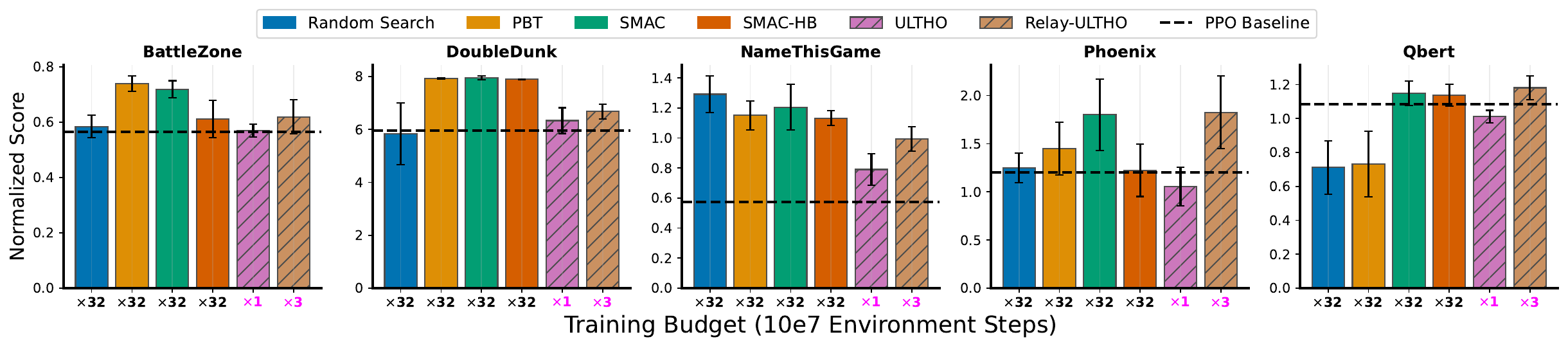}
   \vspace{-20pt}
   \caption{Performance comparison of the two \method algorithms and HPO baselines on the ALE-5 benchmark, in which the mean and standard error are computed using five random seeds. All the scores are normalized using the min-max normalization with the human expert performance. The two ULTHO algorithms achieve similar or even higher performance with a few training budgets as compared to other HPO baselines. An ablation study that uses the same training budgets for all the methods can be found in Appendix~\ref{appendix:ablation}.
   }
   \label{fig:ale_ppo}
   \vspace{-10pt}
\end{figure*}

\begin{figure*}[t!]
  \centering
   \includegraphics[width=\linewidth]{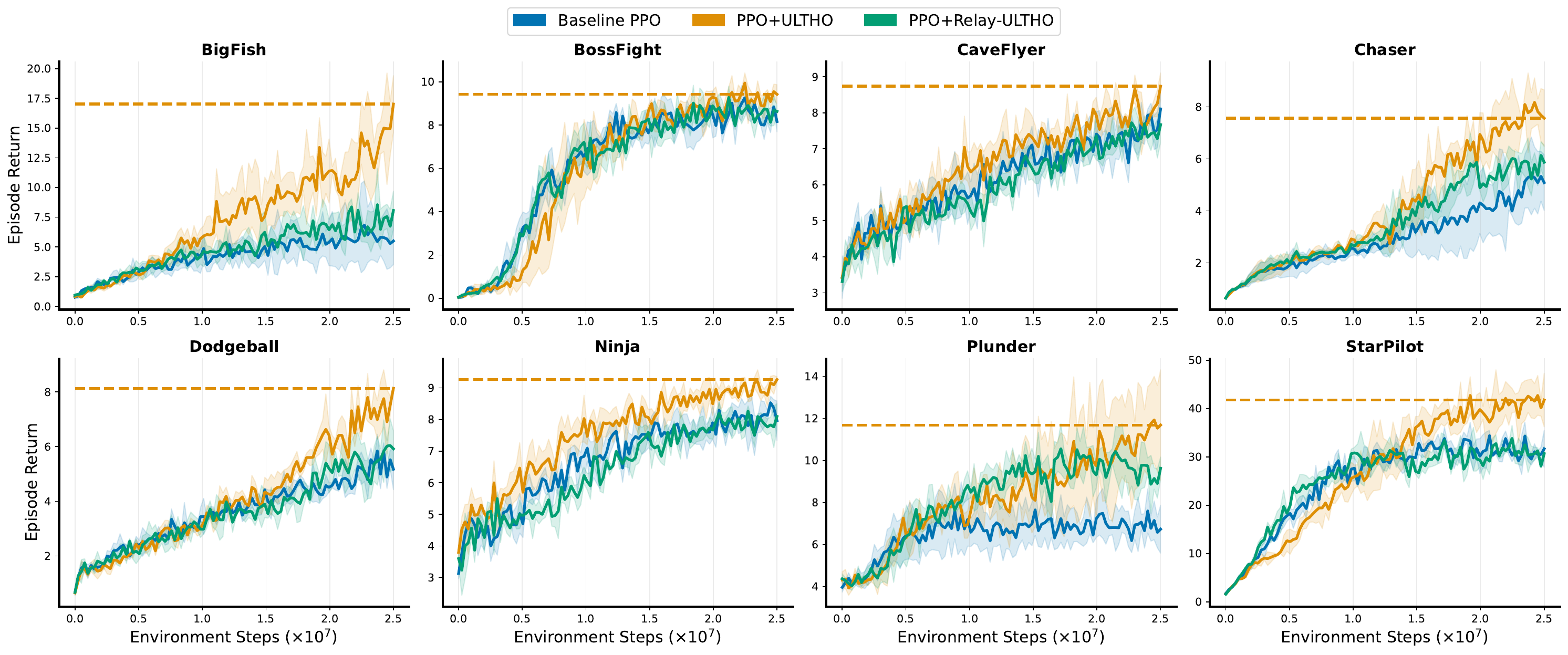}
   \vspace{-20pt}
   \caption{Training performance of the PPO and its combinations with two \method algorithms in eight Procgen environments, in which the mean and standard deviation are computed using five random seeds. ULTHO consistently outperforms the vanilla PPO agent in all the environments, significantly improving the sample efficiency. However, the Relay-ULTHO does not achieve any further performance gains, indicating that a more comprehensive policy, which continuously adapts multiple HPs, is crucial for highly dynamic environments.}
   \label{fig:pg_ppo_8}
   \vspace{-10pt}
\end{figure*}

\subsubsection{HP Clusters} 
We select HP clusters based on the prior practice reported in \cite{jaderberg2017population,paul2019fast,franke2021sampleefficient,huang202237,yuan2025adaptive}. Specifically, our HP clusters include the learning rate (LR), batch size (BS), value loss coefficient (VLC), entropy loss coefficient (ELC), and the number of update epochs (NUE). To ensure a balanced exploration, we let each cluster have the same number of values. For the detailed configuration of each cluster, please refer to Appendix~\ref{appendix:exp}.

\subsubsection{Evaluation Metrics}
For the baselines that require multiple training runs ({\em e.g.}, PBT), we use the maximum observed past scores as their final performance. For our \method algorithm, we simply use the observed episode return at the end of training as its performance. Note that the score of each method on each environment is computed as the average episode returns over 100 episodes and 5 random seeds. Additionally, we compare computational efficiency by evaluating the required training budgets for all the methods.

\subsection{Results Analysis}
The following results analysis is performed based on the pre-defined research questions. We provide the detailed training curves of all the methods and configurations in Appendix~\ref{appendix:curves}.

\subsubsection{Comparison with HPO Baselines}
We first compare the performance of \method and four HPO baselines on the ALE benchmark, and their normalized scores and training budgets are illustrated in Figure~\ref{fig:ale_ppo}. Here, the baseline PPO is trained using the common HPs reported in its original paper \cite{schulman2017proximal}. \method successfully outperforms the PPO baseline in all five environments and achieves the highest performance in two environments, especially in \textit{Q*Bert} environment. In contrast, the SMAC and SMAC+HB rank second and third in terms of average performance across all the environments. The PBT method excels in the \textit{BattleZone} environment, but it fails to outperform the baseline PPO in the \textit{Q*bert} environment. By adaptively tuning HPs within the training run, \method can achieve remarkable efficiency with much lower computational cost as compared to the baselines. 

\begin{table}[h!]
\centering
\begin{tabular}{l|ll}
\toprule
\textbf{Method}       & \textbf{Agg. Mean} & \textbf{Agg. IQM} \\ \bottomrule
PPO          & 40.42$\pm$25.18 & 38.41$\pm$6.59        \\
PBT$\ddagger$         & 33.55$\pm$23.49 & 29.27$\pm$13.68        \\
PB2$\ddagger$         & 44.63$\pm$23.56 & 40.5$\pm$11.41          \\
TMIHF$\ddagger$       & 51.16$\pm$20.16 & 48.7$\pm$4.67          \\
ULTHO        &  \textbf{56.0$\pm$21.87} & \textbf{56.41$\pm$6.95}  \\
Relay-ULTHO & 48.04$\pm$26.88 & 49.01$\pm$5.0                       \\ \bottomrule
\end{tabular}
\caption{Normalized test performance (\%) comparison on the Procgen benchmark. Here, \textbf{IQM} denotes the interquartile mean suggested by \cite{agarwal2021deep}, and $\ddagger$ indicates the training is conducted by a population of agents. Similar to the training performance comparison, the normal ULTHO method achieves the highest score by providing a more comprehensive HPO policy.}
\label{tb:pg_test}
\vspace{-15pt}
\end{table}

Next, we report the performance of the \method on the Procgen benchmark. Similarly, the baseline PPO is trained using the reported HPs in \cite{cobbe2020leveraging}. Figure~\ref{fig:pg_ppo_8} illustrates the training performance comparison in eight environments, and the full curves are provided in Figure~\ref{fig:pg_ppo_16}. \method outperforms the baseline PPO in 15 environments, achieving significant performance gains in environments like \textit{BigFish}, \textit{Chaser}, and \textit{StarPilot}. Additionally, we evaluate \method on the full distribution of levels, and Table~\ref{tb:pg_test} illustrates the performance comparison with three HPO baselines. \method outperforms these baselines regarding both the aggregated mean and IQM. These results demonstrate that \method can effectively select the appropriate HPs at different learning stages and significantly improve learning efficiency even in highly dynamic environments. 


\begin{figure}[h!]
\centering
\vspace{-10pt}
\begin{subfigure}[b]{\linewidth}
    \centering
    \includegraphics[width=\linewidth]{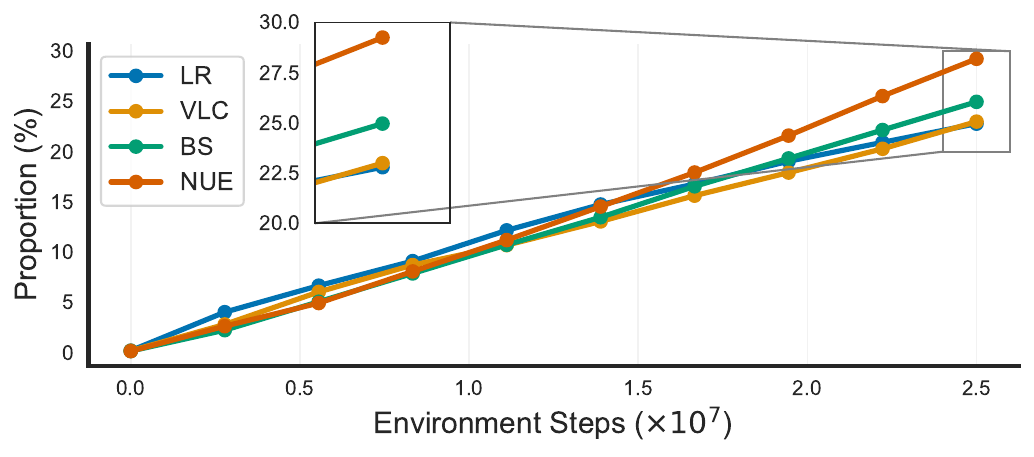}
    \caption{Inter-cluster decision process.}
    \label{fig:pg_ppo_inter_decisions_agg}
\end{subfigure}
\vfill
\begin{subfigure}[b]{\linewidth}
    \centering
    \includegraphics[width=\linewidth]{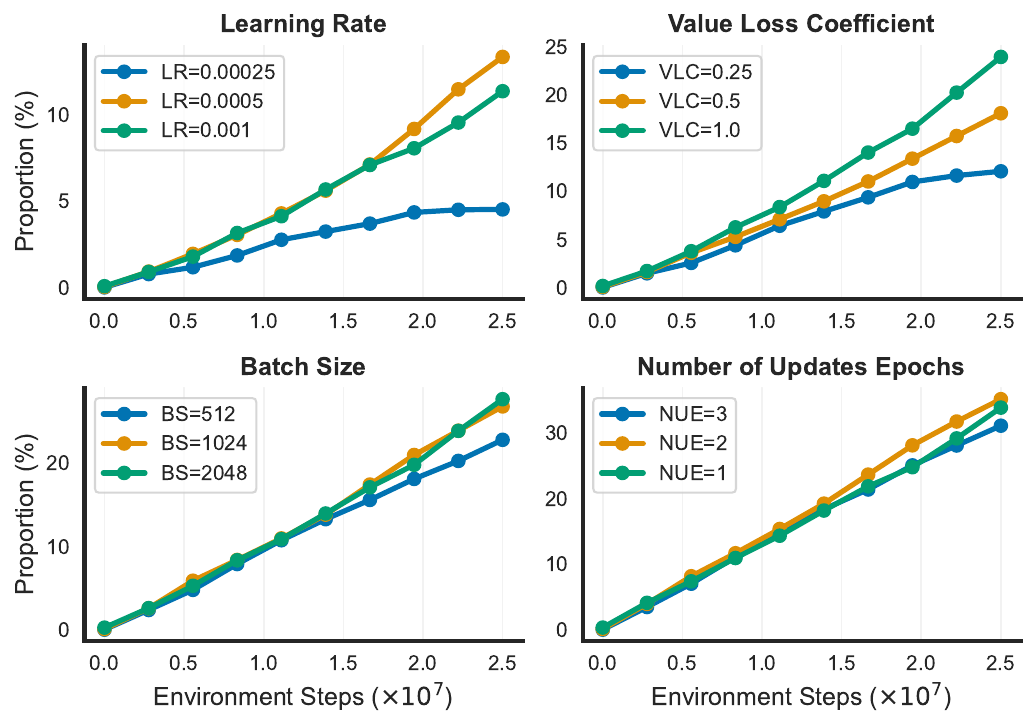}
    \caption{Intra-cluster decision process.}
\end{subfigure}
\vspace{-20pt}
\caption{Aggregated inter-cluster and intra-cluster decision processes of \method on the whole Procgen benchmark. Among the four HP clusters, the NUE cluster is selected the most, suggesting its high impact on overall performance.}
\label{fig:pg_ppo_decision_agg}
\vspace{-10pt}
\end{figure}

\subsubsection{Capability of Relay Optimization}
The high sensitivity to HPs of deep RL makes it risky to optimize multiple HPs simultaneously, leading to the development of the Relay-\method algorithm to enhance exploration and stability. To evaluate its effectiveness, we test it on both the ALE and Procgen benchmarks. Relay-\method produces a much higher performance than the normal version, showing the incremental potential by identifying the COI and NOI from the HP space. However, Relay-\method fails to outperform the normal version on the Procgen benchmark. While Relay-\method is effective in environments with relatively stable dynamics, it struggles in highly non-stationary settings like Procgen, where optimal HP configurations may shift continually across training stages. Therefore, it is feasible to combine the two algorithms to realize a more robust HPO process.

\begin{figure}[t!]
\begin{center}
\includegraphics[width=\linewidth]{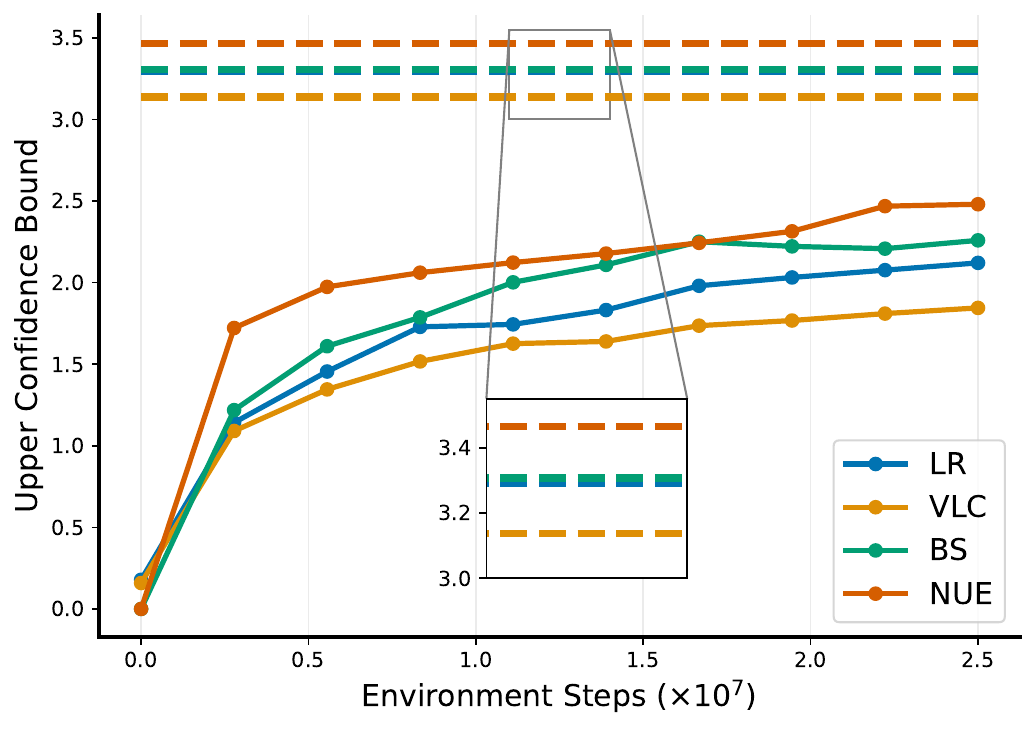}
\vspace{-20pt}
\caption{The variation of aggregated confidence intervals on the whole Procgen environments. Here, the solid line represents the mean value, and the dashed line represents the final upper confidence bound. It is evident that the NUE cluster obtains the highest upper confidence bound at the end of training, which aligns with the selection proportion illustrated in Figure~\ref{fig:pg_ppo_decision_agg}.}
\label{fig:pg_ppo_inter_ct_agg}
\end{center}
\vspace{-20pt}
\end{figure}

\subsubsection{The Detailed Decision Process}
Furthermore, we analyze the detailed decision processes of \nsmethod. Figure~\ref{fig:pg_ppo_inter_decisions_agg} illustrates the cumulative proportion of each cluster selected during the whole training. It is clear that \method primarily selects the NUE cluster, while the other three clusters account for approximately 20\%. For intra-cluster decisions, we find that the LR=5e-4, VLC=1.0, BS=2048, and NUE=2 are the most popular choices. Additionally, Figure~\ref{fig:pg_ppo_inter_ct_agg} illustrates the variation of aggregated confidence intervals of these clusters during the training. \method shows the explicitly distinct preferences of HP clusters, which aligns with the cumulative proportion illustrated in Figure~\ref{fig:pg_ppo_inter_decisions_agg}. These confidence intervals provide a statistical and quantified perspective to identify the importance of each HP cluster, contributing to a more efficient filtering approach. Finally, we provide the detailed decision processes of each method and environment in Appendix~\ref{appendix:decision}.

\begin{figure}[h!]
\begin{center}
\vspace{-10pt}
\centerline{\includegraphics[width=\linewidth]{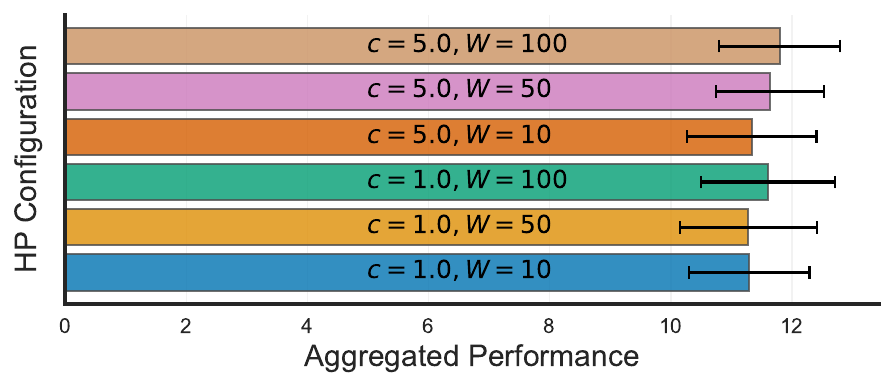}}
\vspace{-10pt}
\caption{Aggregated performance of \method on the Procgen benchmark with different configurations of $c$ and $W$, and the mean and standard error are computed across all the environments. These results demonstrate that ULTHO is robust to the variation of the two internal HPs.}
\label{fig:pg_grid}
\end{center}
\vspace{-30pt}
\end{figure}

\begin{figure*}[t]
\begin{center}
\centerline{\includegraphics[width=\linewidth]{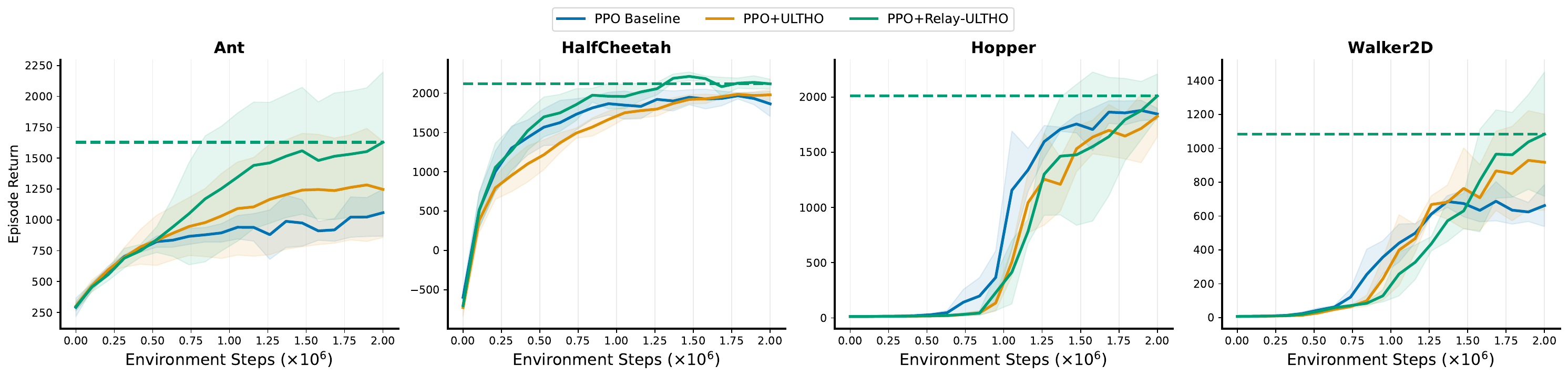}}
\vspace{-10pt}
\caption{Performance of the PPO baseline and two \method algorithms on the PyBullet benchmark, and the mean and standard deviation are computed using five random seeds. Our method can effectively improve the sample efficiency through high-quality HPO in continuous control tasks.}
\label{fig:pb_ppo}
\end{center}
\vspace{-35pt}
\end{figure*}

\subsubsection{Ablation Studies}
While \method achieves efficient HPO, the utilized UCB method relies on two HPs, the exploration coefficient $c$ and the length of the sliding window $W$. To evaluate the robustness of \nsmethod, we perform experiments using different configurations of the two HPs on the Procgen benchmark, and the aggregated training performance is illustrated in Figure~\ref{fig:pg_grid}. It is evident that the \method is relatively insensitive to the choice of $c$ and $W$, yet a bigger $W$ helps achieve a more reliable return estimation. Therefore, \method maintains stable performance across different HPs and can be easily adapted to various training scenarios. More ablation studies regarding the number of HP clusters and training budgets can be found in Appendix~\ref{appendix:ablation}.

\begin{figure}[h]
\begin{center}
\centerline{\includegraphics[width=\linewidth]{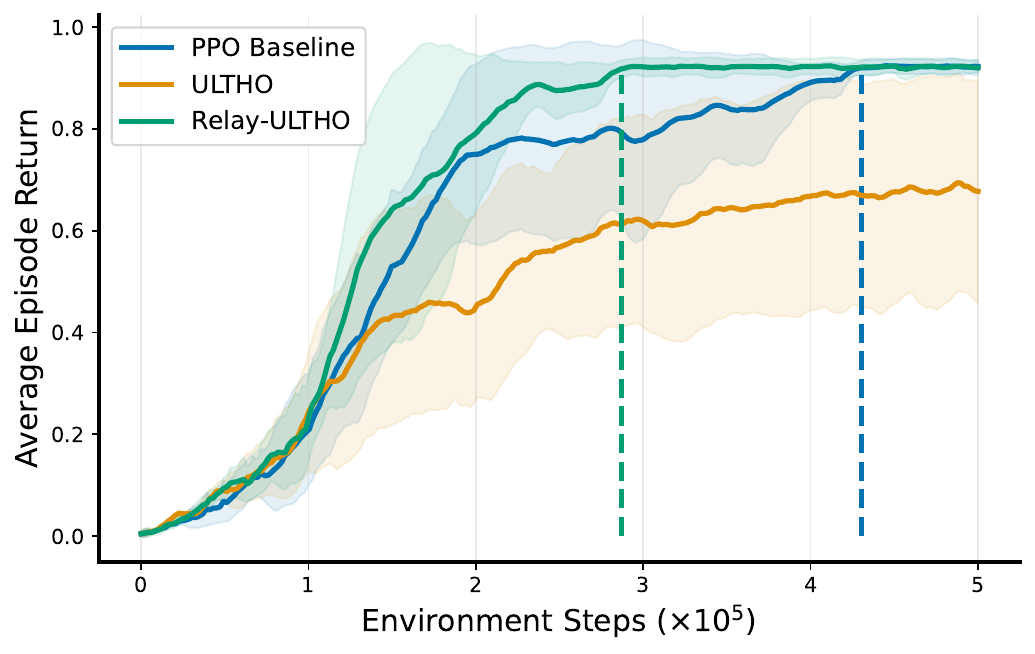}}
\caption{Aggregated performance of \method on the MiniGrid benchmark, and the mean and standard deviation are computed across all the environments. For the relatively stationary environments, Relay-ULTHO can further obtain performance gains through incremental optimization.}
\label{fig:mg_ppo_agg}
\end{center}
\vspace{-30pt}
\end{figure}

\subsubsection{Performance in Sparse-rewards Environments}
Additionally, we evaluate \method on the MiniGrid benchmark with sparse-rewards and goal-oriented environments. Specifically, we conduct experiments using \textit{DoorKey-6$\times$6}, \textit{LavaGapS7}, and \textit{Empty-16$\times$16}. Figure~\ref{fig:mg_ppo_agg} illustrates the aggregated learning curves of the PPO baseline and two \method algorithms. While \method fails to outperform the PPO baseline, the relay optimization takes fewer environment steps to solve the tasks, highlighting its capability to achieve efficient HPO in both dense and sparse-reward settings. More experimental details are provided in Appendix~\ref{appendix:exp}.

\subsubsection{Performance on Continuous Control Tasks}
Finally, we evaluate \method on the PyBullet benchmark with continuous control tasks. Four environments are utilized, namely \textit{Ant}, \textit{HalfCheetah}, \textit{Hopper}, and \textit{Walker2D}. Figure~\ref{fig:pb_ppo} illustrates the aggregated learning curves of the vanilla PPO agent and two \method methods, which produce significant performance gain on the whole benchmark. Similarly, as analyzed before, relay optimization brings advantages in relatively stationary environments. These results underscore the effectiveness of \method in enhancing RL algorithms across both discrete and continuous control tasks. Additional experimental details can be found in Appendix~\ref{appendix:exp}.

\section{Discussion}
In this paper, we investigated the problem of HPO in deep RL and proposed an ultra-lightweight yet powerful framework entitled \nsmethod, which performs HPO on the fly and requires no complex learning process. \method formulates the HPO process as a multi-armed bandit with clustered arms, enabling efficient and adaptive HP selection across different learning stages. We evaluate \method on ALE, Procgen, MiniGrid, and PyBullet benchmarks. Extensive experiments demonstrate that \method can effectively enhance RL algorithms with a simple architecture, contributing to the development of advanced and automated RL systems.

Still, there are currently remaining limitations to this work. As a lightweight framework, \method currently operates on categorical HP values, which discretizes HP search into predefined clusters. While this ensures computational efficiency and stable optimization, it may limit the granularity of HP tuning in some cases. Extending \method to handle continuous HP spaces for more precise tuning is a promising direction for future work. Additionally, while \method uses a UCB-based strategy for exploration-exploitation balance, we have yet to explore alternative bandit algorithms, such as Exp3 or Thompson sampling. Evaluating these strategies within \method could provide valuable insights into their impact on RL-specific HPO and potentially further improve performance in dynamic environments. Future work will focus on mitigating these issues and further enhancing \nsmethod, making it more robust, adaptive, and efficient for HPO in deep RL.

\clearpage\newpage

\section*{Acknowledgements}
This work is supported, in part, by the HKSAR RGC under Grant No. PolyU 15224823, the Guangdong Basic and Applied Basic Research Foundation under Grant No. 2024A1515011524, the NSFC under Grant No. 62302246, the ZJNSFC under Grant No. LQ23F010008, and the Ningbo under Grants No. 2023Z237 \& 2024Z284 \& 2024Z289 \& 2023CX050011 \& 2025Z038. We thank the high-performance computing center at Eastern Institute of Technology and Ningbo Institute of Digital Twin for providing the computing resources.

{
    \small
    \bibliographystyle{ieeenat_fullname}
    \bibliography{main}
}

\clearpage\newpage

\appendix
\onecolumn
\section{Algorithmic Baselines}\label{appendix:baseline}
\subsection{PPO}
Proximal policy optimization (PPO) \cite{schulman2017proximal} is an on-policy algorithm that is designed to improve the stability and sample efficiency of policy gradient methods, which uses a clipped surrogate objective function to avoid large policy updates. 

The policy loss is defined as:
\begin{equation}
    L_{\pi}(\bm{\theta})=-\mathbb{E}_{\tau\sim\pi}\left[\min\left(\rho_{t}(\bm{\theta})A_{t},{\rm clip}\left(\rho_{t}(\bm{\theta}),1-\epsilon,1+\epsilon\right)A_{t}\right)\right],
\end{equation}
where 
\begin{equation}
    \rho_{t}(\bm{\theta})=\frac{\pi_{\bm\theta}(\bm{a}_{t}|\bm{s}_{t})}{\pi_{\bm\theta_{\rm old}}(\bm{a}_{t}|\bm{s}_{t})},
\end{equation}
and $\epsilon$ is a clipping range coefficient.

Meanwhile, the value network is trained to minimize the error between the predicted return and a target of discounted returns computed with generalized advantage estimation (GAE) \cite{schulman2015high}:
\begin{equation}
    L_{V}(\bm{\phi})=\mathbb{E}_{\tau\sim\pi}\left[\left(V_{\bm\phi}(\bm{s})-V_{t}^{\rm target}\right)^{2}\right].
\end{equation}

\subsection{Random Search}
Random search (RS) \cite{bergstra2012random} is a simple yet effective method for hyperparameter optimization that randomly samples from the configuration space instead of exhaustively searching through all combinations. Compared to grid search, RS is particularly efficient in high-dimensional spaces, where it can outperform grid search by focusing on a wider area of the search space. It is especially beneficial when only a few hyperparameters significantly influence the model's performance, as it can effectively explore these critical dimensions without the computational cost of grid search. RS is also highly parallelizable and flexible, allowing for dynamic adjustments to the search process.

\subsection{PBT}
Population-based training (PBT) \cite{jaderberg2017population} is an asynchronous optimization method designed to optimize both model parameters and hyperparameters simultaneously. Unlike traditional hyperparameter tuning methods that rely on fixed schedules for hyperparameters, PBT adapts hyperparameters during training by exploiting the best-performing models and exploring new hyperparameter configurations. PBT operates by maintaining a population of models and periodically evaluating their performance, and using this information to guide the optimization of hyperparameters and model weights. This approach ensures efficient use of computational resources while achieving faster convergence and improved final performance, particularly in RL and generative modeling tasks. 

\subsection{PB2}
Population-based bandits (PB2) \cite{hazan2019provably} enhances PBT by using a multi-armed bandit approach to dynamically select and optimize hyperparameters based on performance. This method improves the exploration-exploitation trade-off, allocating resources to the most promising configurations and reducing computational costs while accelerating convergence.

\subsection{SMAC+HB}
SAMC \cite{lindauer2022smac3} is a powerful framework for hyperparameter optimization that leverages Bayesian optimization (BO) to efficiently find well-performing configurations. It uses a random forest model as a surrogate to predict the performance of hyperparameter configurations, which is particularly effective for high-dimensional spaces. SMAC optimizes the hyperparameters of machine learning algorithms by iteratively selecting configurations based on a probabilistic model of the objective function. Additionally, SMAC integrates with Hyperband \cite{li2018hyperband} for more efficient resource allocation.

\subsection{TMIHF}
TMIHF \cite{parker2021tuning} introduces a novel approach for optimizing both continuous and categorical hyperparameters in reinforcement learning (RL). This method builds on the population-based bandits (PB2) \cite{parker2020provably} framework and addresses its limitation of only handling continuous hyperparameters. By employing a time-varying multi-armed bandit algorithm, TMIHF efficiently selects both continuous and categorical hyperparameters in a population-based training setup, thereby improving sample efficiency and overall performance. The algorithm's hierarchical structure allows it to model the dependency between categorical and continuous hyperparameters, which is crucial for tasks like data augmentation in RL environments.

\clearpage\newpage

\section{Experimental Setting}\label{appendix:exp}
\subsection{Arcade Learning Environment}
\begin{figure*}[h!]
\begin{center}
\centerline{\includegraphics[width=0.7\linewidth]{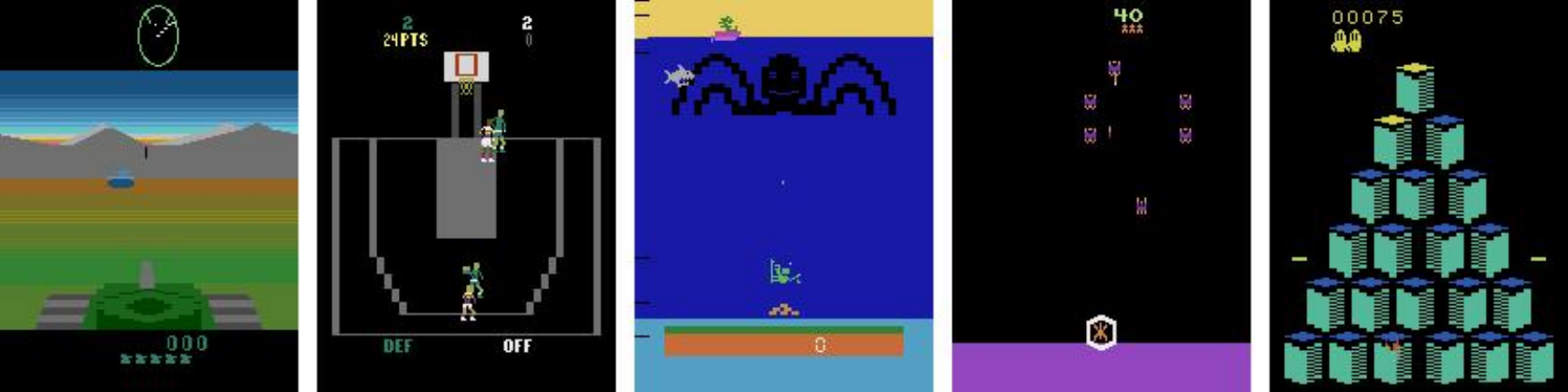}}
\caption{Screenshots of the ALE-5 environments. From left to right: \textit{BattleZone}, \textit{DoubleDunk}, \textit{NameThisGame}, \textit{Phoenix}, and \textit{Q*Bert}.}
\label{fig:ale_screenshots}
\end{center}
\end{figure*}

\noindent\textbf{PPO+\nsmethod}. In this part, we utilize the implementation of \cite{pytorchrl} for the PPO algorithm, and train the agent for 10M environment steps in each environment. For the hyperparameter clusters, we select the batch size, value loss coefficient, and entropy loss coefficient as the candidates, and the detailed values of each cluster are listed in Table~\ref{tb:hp_clusters}. Additionally, we run a grid search over the exploration coefficient $c\in\{1.0, 5.0\}$ and the size of the sliding window used to compute the $Q$-values $W\in\{10, 50, 100\}$ to study the robustness of \nsmethod. Finally, Table~\ref{tb:ppo_hp} illustrates the PPO hyperparameters, which remain fixed throughout all the experiments except for the hyperparameter clusters. 

\hspace*{\fill}

\noindent\textbf{PPO+Relay-\nsmethod}. At the end of the \textbf{PPO+\nsmethod} experiments, we count the number of times each cluster is selected and find out the cluster of interest and the neglected cluster of interest. Then we perform the experiments with the two clusters separately before reporting the best-performing cluster. Therefore, the actual training budget of Relay-\method is three times that of \nsmethod, {\em i.e.}, 30M environment steps. Similarly, we run a grid search over the exploration coefficient and the sliding window size and report the best results.

\hspace*{\fill}

\noindent\textbf{HPO Baselines}. For RS, PBT, SMAF, and SMAF+HB, we utilize the implementations provided in ARLBench \cite{becktepe2024arlbench}, which is a benchmark for hyperparameter optimization in RL and allows comparisons of diverse approaches. The training budget for each method is 320M environment steps with five runs, in which each configuration is evaluated on three random seeds. The results reported in this paper are directly obtained from the provided dataset in the ARLBench.

\subsection{Procgen}
\begin{figure*}[h!]
\begin{center}
\centerline{\includegraphics[width=\linewidth]{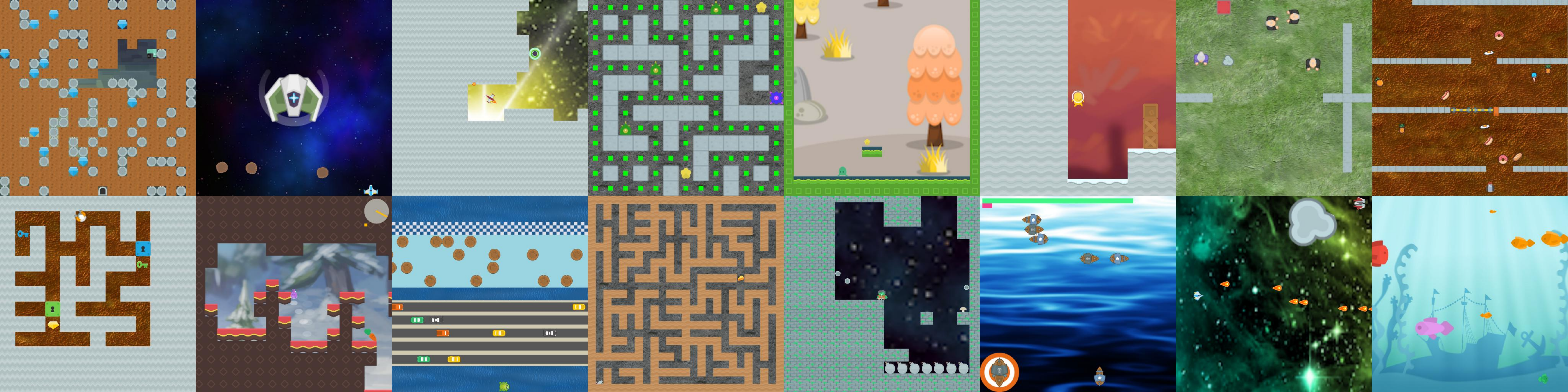}}
\caption{Screenshots of the sixteen Procgen environments.}
\label{fig:procgen_screenshots}
\end{center}
\end{figure*}

\noindent\textbf{PPO+\nsmethod}. In this part, we utilize the implementation of CleanRL for the PPO algorithm and train the agent for 25M environment steps on 200 levels before testing it on the full distribution of levels. For the hyperparameter clusters, we select the batch size, value loss coefficient, entropy loss coefficient, and number of update epochs as the candidates, and the detailed values of each cluster are listed in Table~\ref{tb:hp_clusters}. Similarly, we run a grid search over the exploration coefficient $c\in\{1.0, 5.0\}$ and the size of the sliding window used to compute the $Q$-values $W\in\{10, 50, 100\}$ to study the robustness of \nsmethod. Finally, Table~\ref{tb:ppo_hp} illustrates the PPO hyperparameters, which remain fixed throughout all the experiments except for the hyperparameter clusters. 

\hspace*{\fill}

\noindent\textbf{PPO+Relay-\nsmethod}. Similar to the ALE experiments, we identify the two clusters of interest at the end of the \method experiments. Then we also perform the experiments with the two clusters separately before reporting the best-performing cluster. The actual training budget of Relay-\method is 75M environment steps for Procgen. Finally, we run a grid search over the exploration coefficient and the sliding window size and report the best results.

\hspace*{\fill}

\noindent\textbf{HPO Baselines}. For PBT, PB2, and TMIHF, we leverage the official implementations reported in \cite{parker2021tuning}. For each method, the population size is set as 4, and each is trained for 25M environment steps. Therefore, the total training budget is 100M environment steps. The results reported in this paper are directly obtained from the \cite{parker2021tuning}.


\subsection{MiniGrid}
\begin{figure*}[h!]
\begin{center}
\centerline{\includegraphics[width=0.7\linewidth]{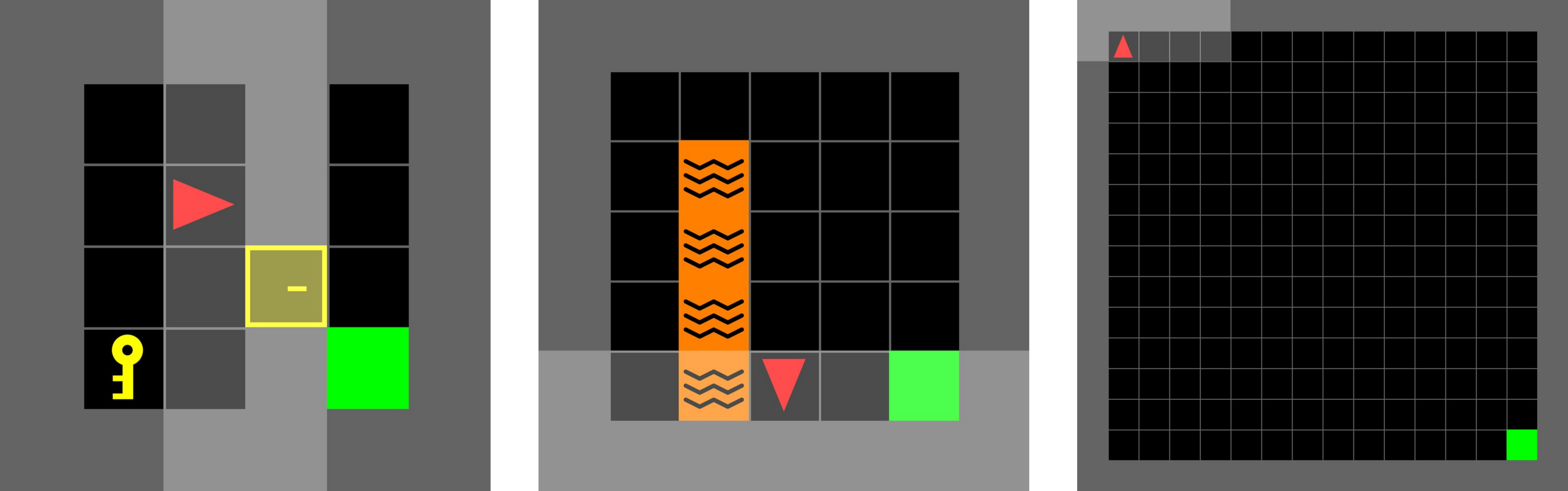}}
\caption{Screenshots of the three MiniGrid environments. From left to right: \textit{DoorKey-6$\times$6}, \textit{LavaGapS7}, and \textit{Empty-16$\times$16}.}
\label{fig:minigrid_screenshots}
\end{center}
\end{figure*}

In this part, we use the implementation of \cite{MinigridMiniworld23} for the PPO agent and train each agent for 500K environment steps. For the hyperparameter clusters, we select the learning rate, batch size, and value loss coefficient as the candidates, and the detailed values of each cluster are listed in Table~\ref{tb:hp_clusters}. The experiment workflow of \method and Relay-\method is the same as the experiments above. Finally, Table~\ref{tb:ppo_hp} illustrates the PPO hyperparameters, which remain fixed throughout all the experiments except for the hyperparameter clusters. 

\subsection{PyBullet}
\begin{figure*}[h!]
\begin{center}
\centerline{\includegraphics[width=0.75\linewidth]{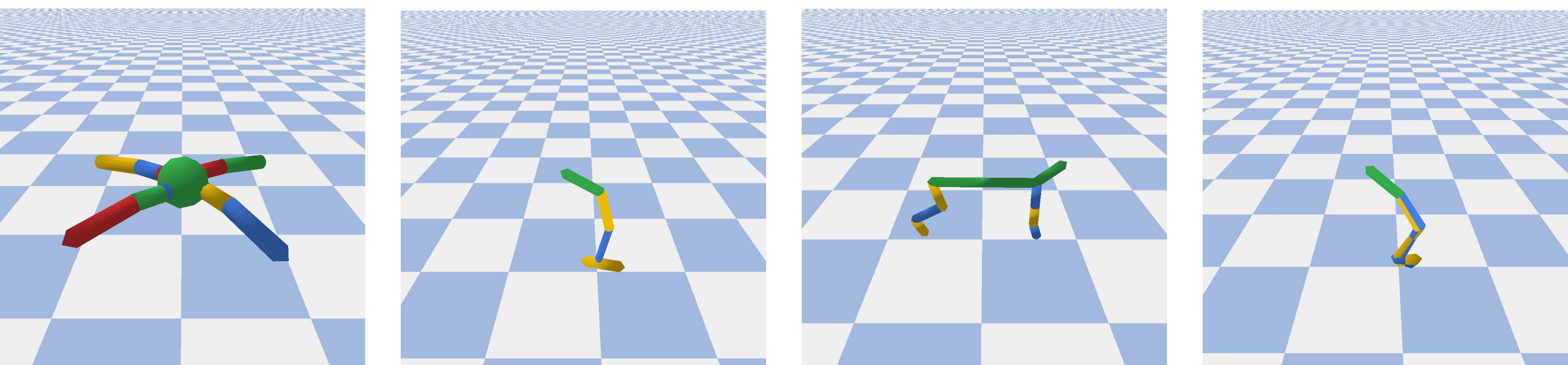}}
\caption{Screenshots of the four PyBullet environments. From left to right: \textit{Ant}, \textit{Hopper}, \textit{HalfCheetah}, and \textit{Walker2D}.}
\label{fig:pybullet_screenshots}
\end{center}
\end{figure*}

Finally, we perform the experiments on the PyBullet benchmark using the PPO implementation of \cite{pytorchrl}, and train each agent for 2M environment steps. Here, we leverage state-based observation rather than image-based observations. For the hyperparameter clusters, we select the learning rate, batch size, and value loss coefficient as the candidates, and the detailed values of each cluster are listed in Table~\ref{tb:hp_clusters}. We also run experiments for both \method and Relay-\method algorithms and report the best results. Likely, Table~\ref{tb:ppo_hp} illustrates the PPO hyperparameters, which remain fixed throughout all the experiments except for the hyperparameter clusters. 

\begin{table}[h!]
\centering
\begin{tabular}{l|llll}
\toprule
\textbf{HP Cluster}      & \textbf{ALE} & \textbf{Procgen} & \textbf{MiniGrid} & \textbf{PyBullet} \\ \midrule
Learning Rate            & N/A          & \{2.5e-4, 5e-4, 1e-3\}           & \{1e-3, 2.5e-3, 5e-3\}            & \{2e-4, 5e-4, 7e-4\}            \\
Batch Size               & \{128, 256, 512\}      & \{512, 1024, 2048\}           & \{128, 256, 512\}            & \{64, 128, 256\}            \\
Vale Loss Coefficient    & \{0.25, 0.5, 1.0\}       & \{0.25, 0.5, 1.0\}           & \{0.25, 0.5, 1.0\}            & \{0.25, 0.5, 1.0\}            \\
Entropy Loss Coefficient & \{0.01, 0.05, 0.1\}       & N/A              & N/A               & N/A               \\
Number of Update Epochs  & N/A          & \{3, 2, 1\}           & N/A               & N/A               \\ \bottomrule
\end{tabular}
\caption{The selected hyperparameter clusters for each benchmark.}
\label{tb:hp_clusters}
\end{table}

\begin{table}[h!]
\centering
\begin{tabular}{lllll}
\toprule
\textbf{Hyperparameter}   & \textbf{ALE} & \textbf{Procgen}           & \textbf{MiniGrid} & \textbf{PyBullet}   \\ \midrule
Observation downsampling   & (84, 84)     & (64, 64, 3)      & (7,7,3)           & N/A               \\
Observation normalization  & / 255.       & / 255.           & No                & Yes               \\
Reward normalization       & Yes          & Yes              & No                & Yes               \\
LSTM                       & No           & No               & No                & No                \\
Stacked frames             & 4            & No               & No                & N/A               \\
Environment steps          & 10000000     & 25000000         & 500000            & 2000000           \\
Episode steps              & 128          & 256              & 128               & 2048              \\
Number of workers          & 1            & 1                & 1                 & 1                 \\
Environments per worker    & 8            & 64               & 16                & 1                 \\
Optimizer                  & Adam         & Adam             & Adam              & Adam              \\
Learning rate              & 2.5e-4       & 5e-4             & 1e-3              & 2e-4              \\
GAE coefficient            & 0.95         & 0.95             & 0.95              & 0.95              \\
Action entropy coefficient & 0.01         & 0.01             & 0.01              & 0                 \\
Value loss coefficient     & 0.5          & 0.5              & 0.5               & 0.5               \\
Value clip range           & 0.2          & 0.2              & 0.2               & N/A               \\
Max gradient norm          & 0.5          & 0.5              & 0.5               & 0.5               \\
Batch size                 & 256          & 2048             & 256               & 64                \\
Discount factor            & 0.99         & 0.999            & 0.99              & 0.99              \\ \bottomrule
\end{tabular}
\caption{The PPO hyperparameters for the four benchmarks. These remain fixed for all experiments except for the selected clusters.}
\label{tb:ppo_hp}
\end{table}

\clearpage\newpage

\section{Learning Curves}\label{appendix:curves}

\begin{figure*}[h!]
\begin{center}
\centerline{\includegraphics[width=\linewidth]{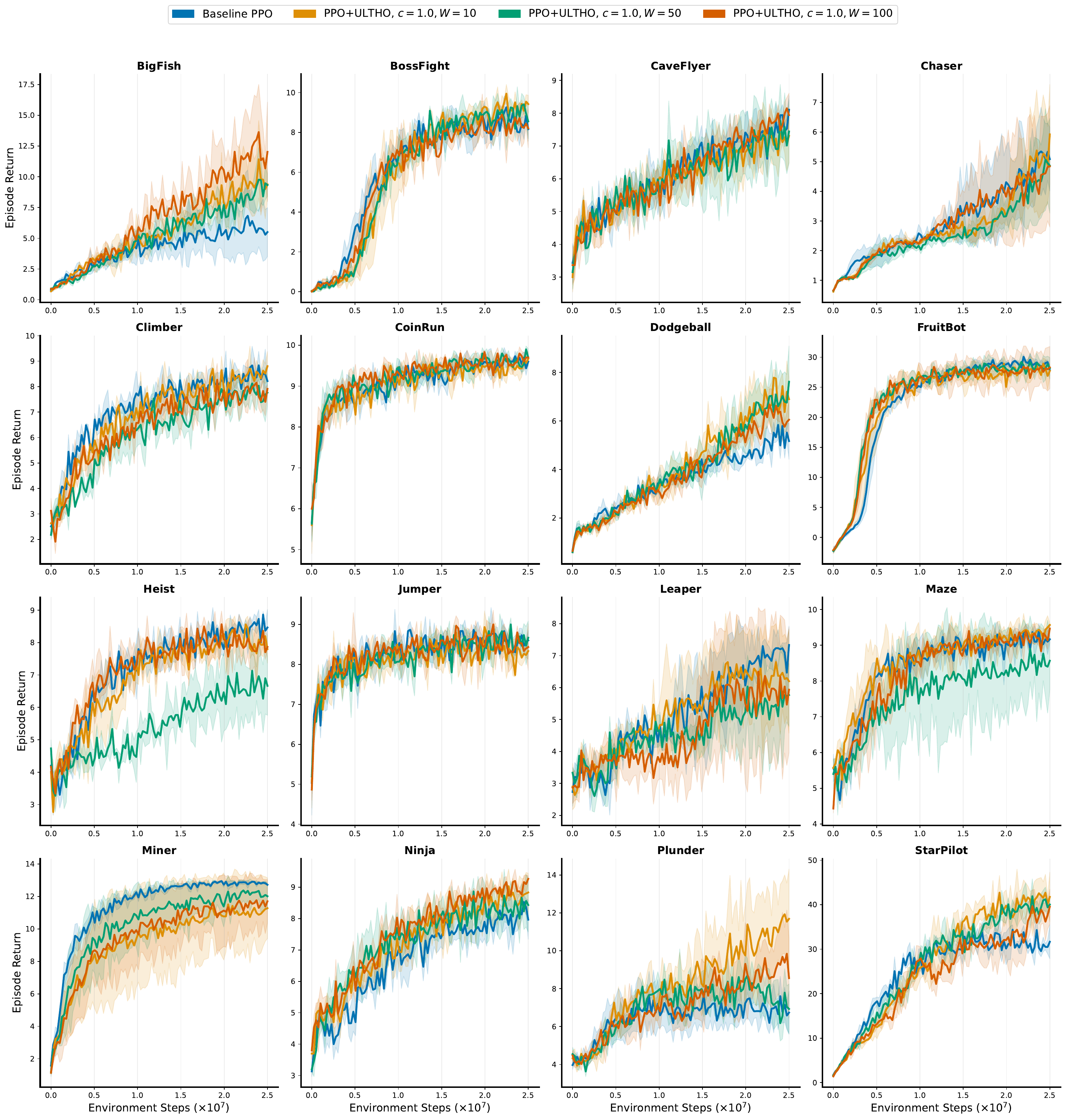}}
\caption{Learning curves of the vanilla PPO agent and \method with different sizes of the sliding window on the Procgen benchmark. Here, the exploration coefficient $c$ is set as 1.0. The mean and standard deviation are computed over five runs with different seeds.}
\label{fig:pg_ultho_ucb_window_c=1.0}
\end{center}
\end{figure*}

\begin{figure*}[h!]
\begin{center}
\centerline{\includegraphics[width=\linewidth]{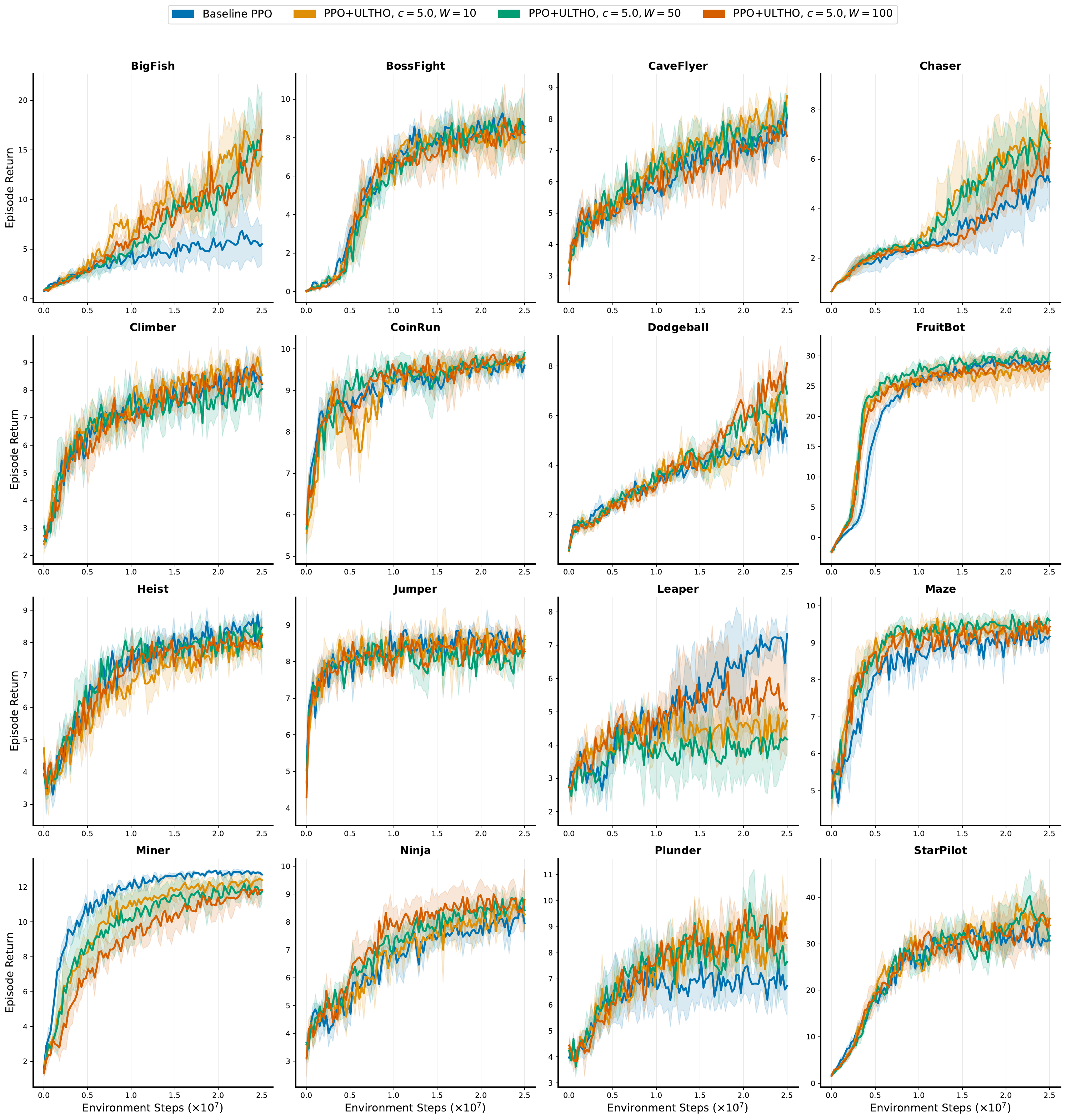}}
\caption{Learning curves of the vanilla PPO agent and \method with different sizes of the sliding window on the Procgen benchmark. Here, the exploration coefficient $c$ is set as 5.0. The mean and standard deviation are computed over five runs with different seeds.}
\label{fig:pg_ultho_ucb_window_c=5.0}
\end{center}
\end{figure*}

\begin{figure*}[h!]
\begin{center}
\centerline{\includegraphics[width=\linewidth]{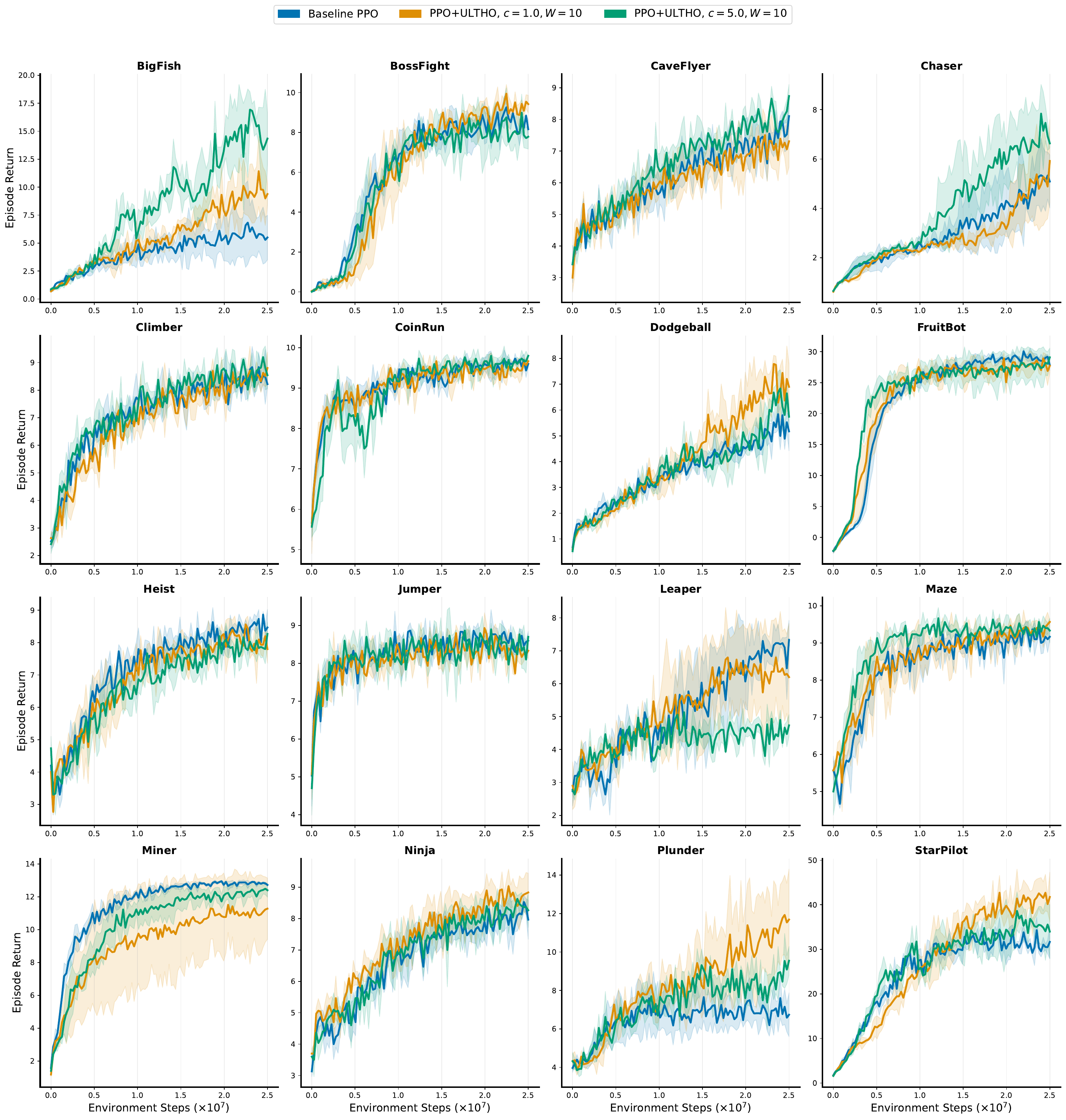}}
\caption{Learning curves of the vanilla PPO agent and \method with different exploration coefficients on the Procgen benchmark. Here, the size $W$ of the sliding window is set as 10. The mean and standard deviation are computed over five runs with different seeds.}
\label{fig:pg_ultho_ucb_expl_w=10}
\end{center}
\end{figure*}

\begin{figure*}[h!]
\begin{center}
\centerline{\includegraphics[width=\linewidth]{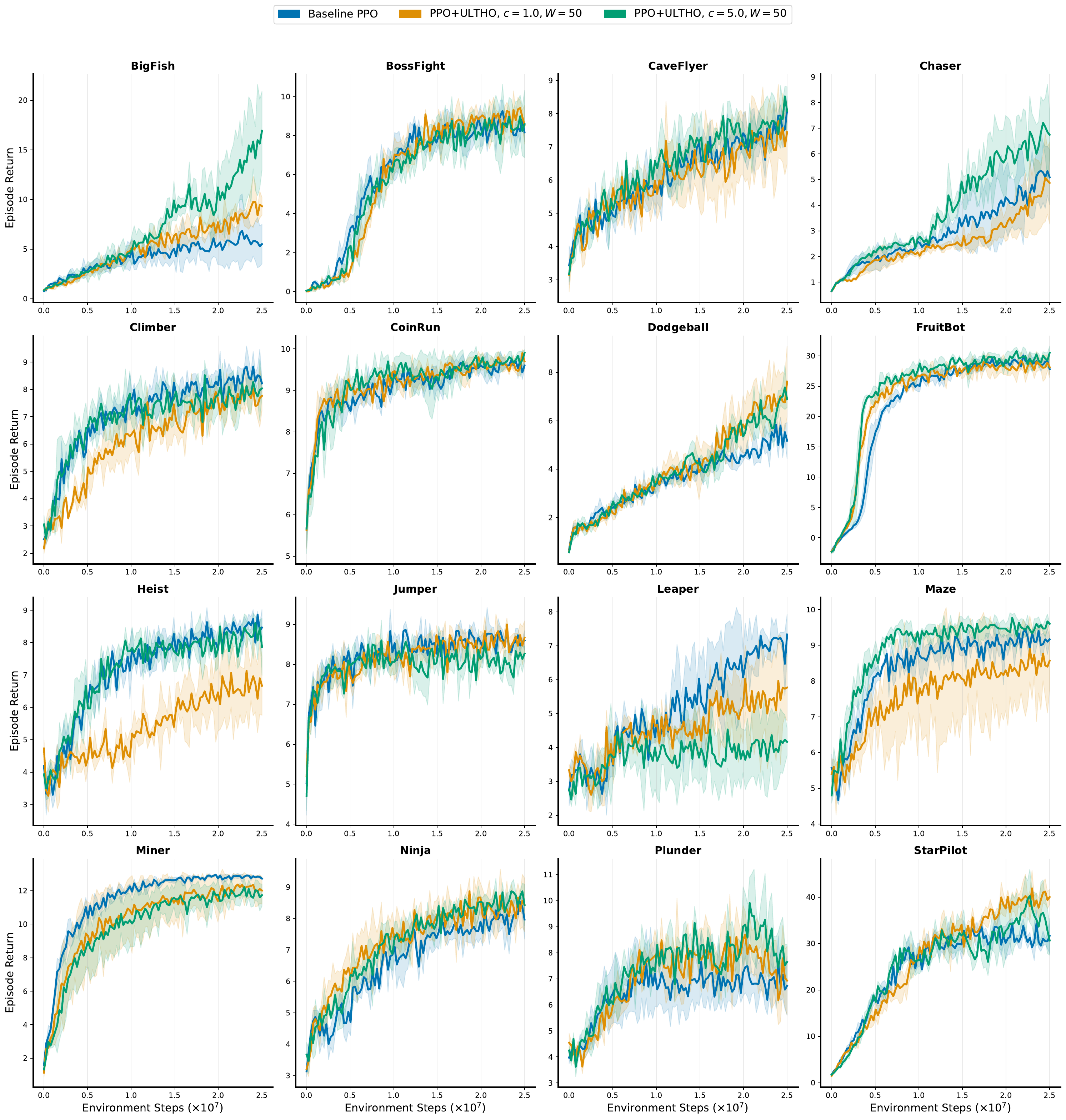}}
\caption{Learning curves of the vanilla PPO agent and \method with different exploration coefficients on the Procgen benchmark. Here, the size $W$ of the sliding window is set as 50. The mean and standard deviation are computed over five runs with different seeds.}
\label{fig:pg_ultho_ucb_expl_w=50}
\end{center}
\end{figure*}

\begin{figure*}[h!]
\begin{center}
\centerline{\includegraphics[width=\linewidth]{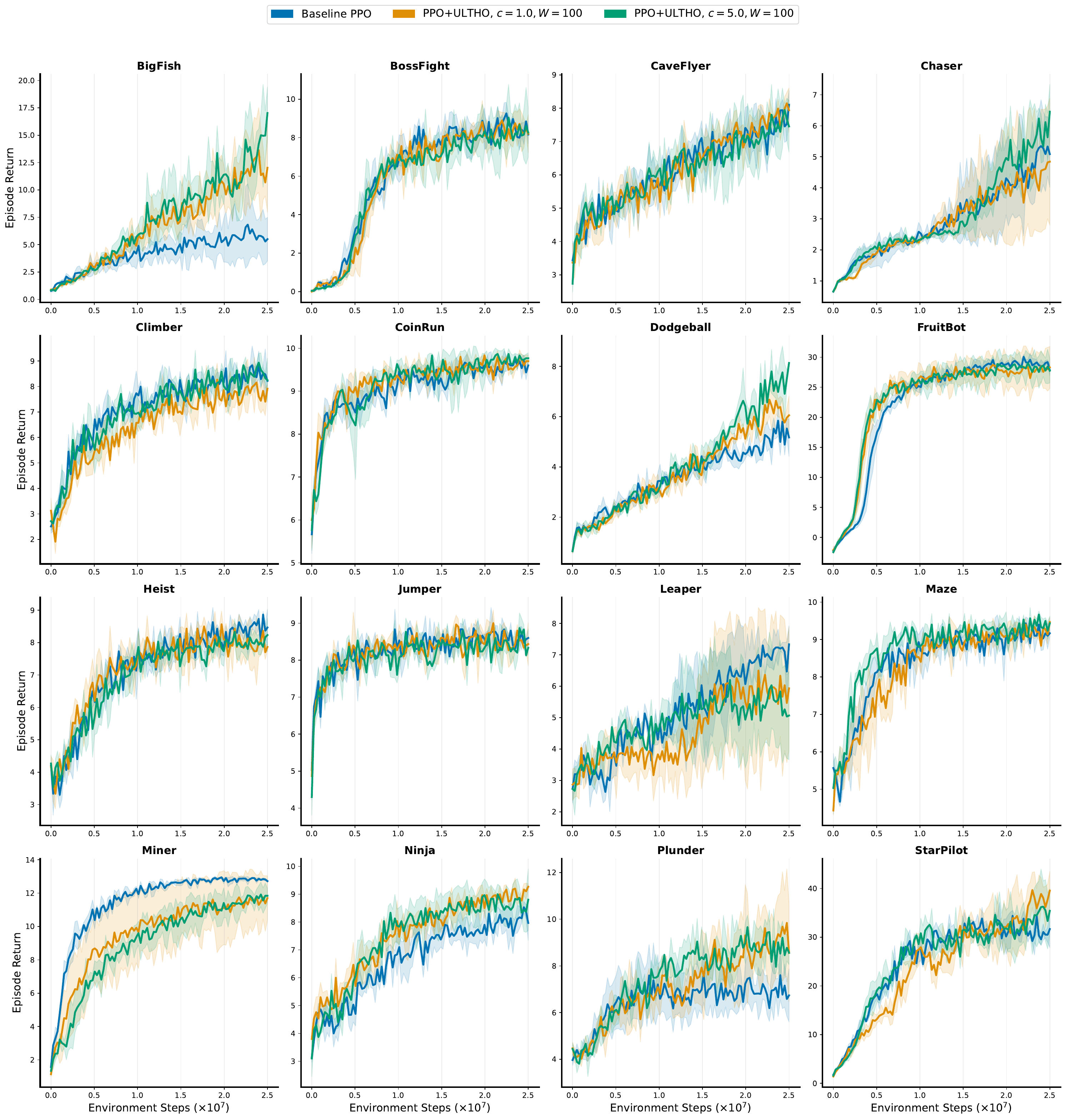}}
\caption{Learning curves of the vanilla PPO agent and \method with different exploration coefficients on the Procgen benchmark. Here, the size $W$ of the sliding window is set as 100. The mean and standard deviation are computed over five runs with different seeds.}
\label{fig:pg_ultho_ucb_expl_w=100}
\end{center}
\end{figure*}

\begin{figure*}[h!]
\begin{center}
\centerline{\includegraphics[width=\linewidth]{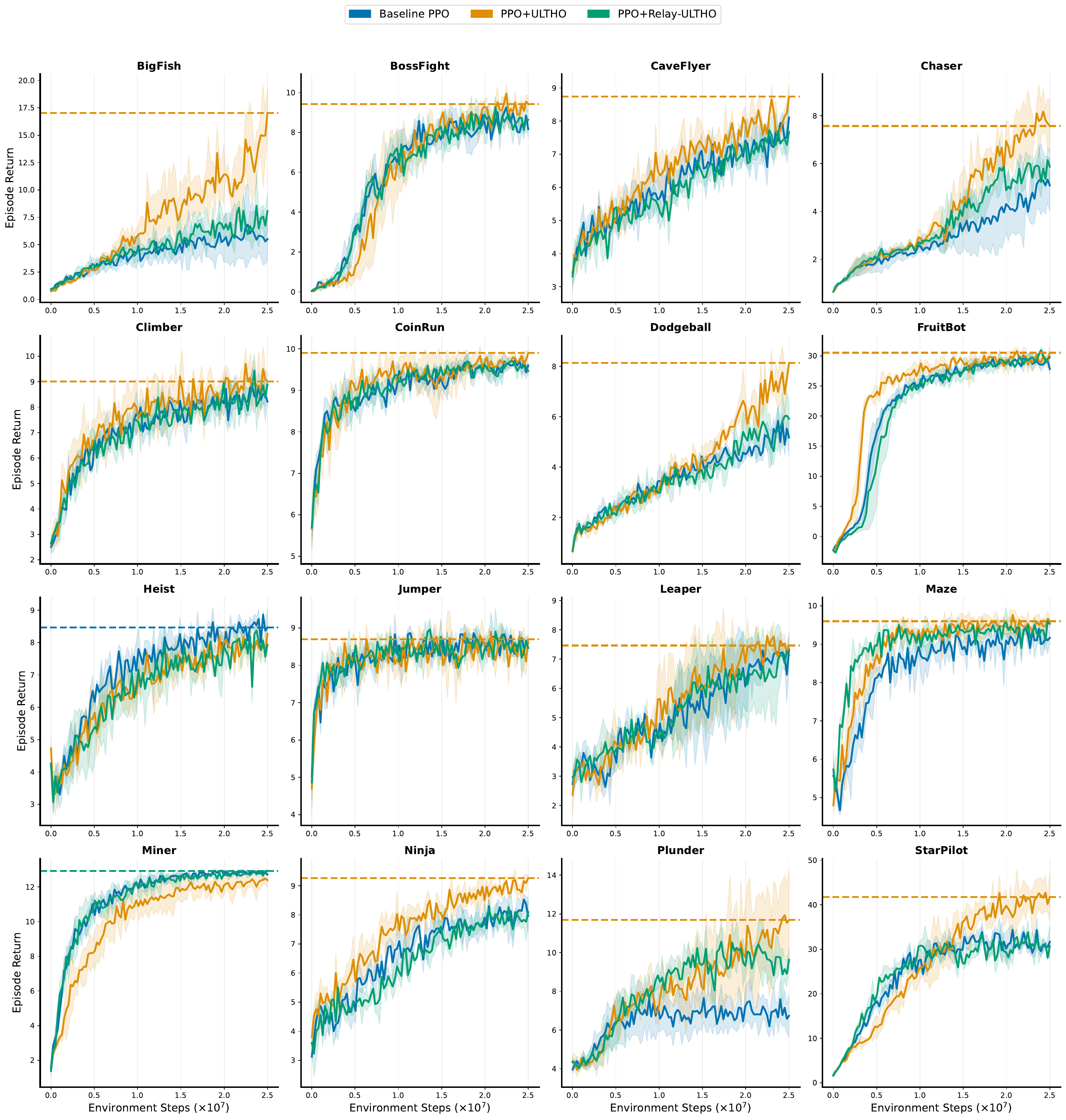}}
\caption{Learning curves of the vanilla PPO agent and two \method algorithms on the Procgen benchmark. The mean and standard deviation are computed over five runs with different seeds.}
\label{fig:pg_ppo_16}
\end{center}
\end{figure*}

\clearpage\newpage

\section{Ablation Studies}\label{appendix:ablation}

\noindent\textbf{ULTHO versus baselines (same training budgets)}. The Figure below compares ULTHO and the baselines on the ALE-5 benchmark using the same training budget, 100M environment steps. By achieving highly frequent HP tuning within a limited budget, ULTHO consistently outperforms the baselines.
\begin{figure*}[h!]
\begin{center}
\centerline{\includegraphics[width=0.4\linewidth]{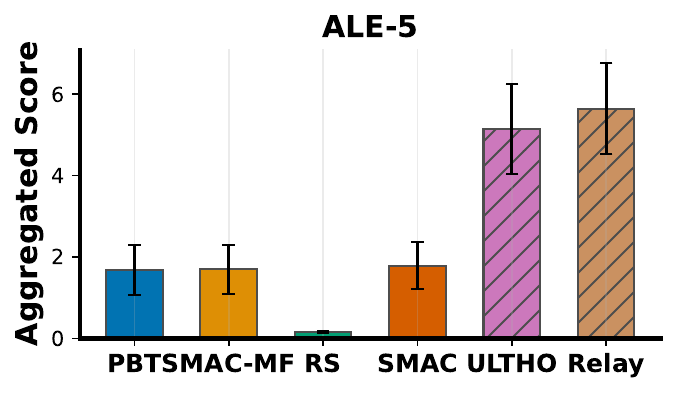}}
\caption{The aggregated performance comparison of ULTHO and baselines on the ALE-5 benchmark with the same training budgets.}
\label{fig:ale_100m}
\end{center}

\end{figure*}

\vspace{-25pt}

\noindent\textbf{ULTHO's performance with varying sets of HPs}. The Figure below compares the performance of ULTHO on the Procgen benchmark with the varying number of HP clusters and intra-cluster HPs. There is a notable performance gain when using more HP clusters. However, as a lightweight framework and constrained by the capability of MAB algorithms, continually increasing the types and granularity of HPs only achieves limited performance gains. 
\begin{figure}[h!]
\begin{minipage}{0.38\linewidth} 
    \centering
    \includegraphics[width=\linewidth]{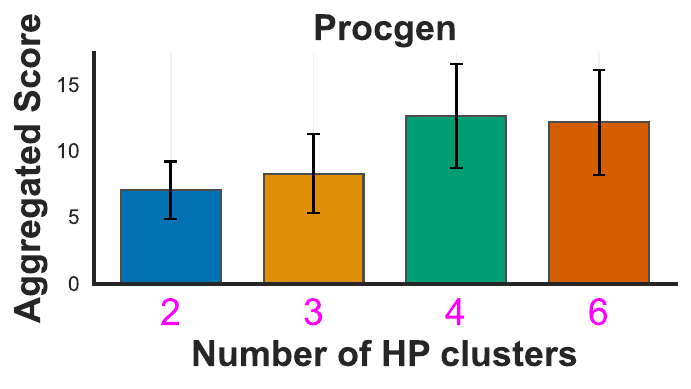}
\label{fig:pg_varing_hpc}
\end{minipage}
\begin{minipage}{0.58\linewidth} 
    \centering
    \includegraphics[width=\linewidth]{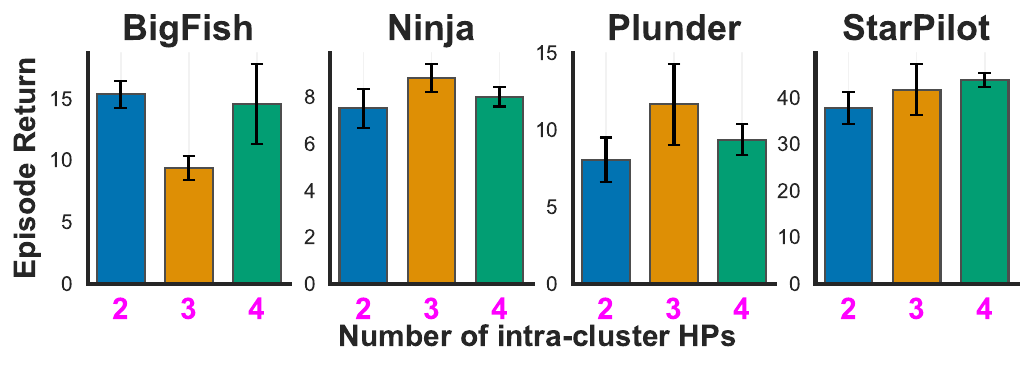}
\label{fig:pg_varing_ihpc}
\end{minipage}
\vspace{-20pt}
\caption{ULTHO's performance comparison with the varying number of HP clusters and intra-cluster HPs.}
\label{fig:varying_hps}
\end{figure}


\noindent\textbf{ULTHO's performance with off-policy RL algorithms}. It is simple to apply ULTHO to off-policy RL algorithms. Take SAC \cite{haarnoja2018soft} for example, every $T$ steps, we use ULTHO to select a HP before using it for model update in the next $T$ steps. The utility of each HP can be evaluated using the action value function: $U_{i}(\bm{\psi})=\frac{1}{W}\sum_{j=1}^{W}\frac{1}{|\mathcal{B}_j|}\sum_{(\bm{s},\bm{a})\sim\mathcal{B}_j}Q(\bm{s},\bm{a})$, where $\mathcal{B}$ is a sampled batch. We perform an experiment using two tasks from the DMC benchmark \cite{tassa2018deepmind}, and the Figure above indicates that ULTHO effectively enhances SAC's performance. However, at the current stage, we do not recommend applying ULTHO to off-policy algorithms as the return estimation based on sampling from a replay buffer is relatively unstable.

\begin{figure}[h!]
    \centering
    \includegraphics[width=0.8\linewidth]{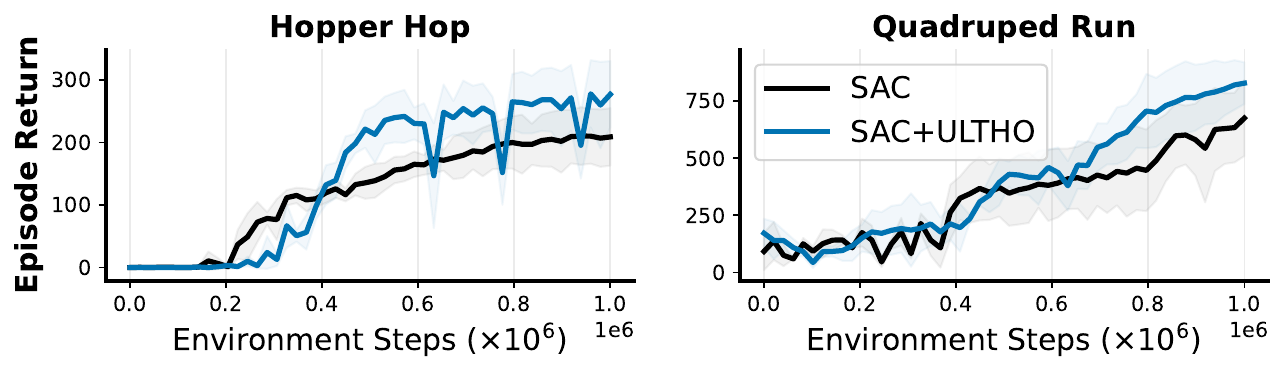}
    \caption{The performance of SAC+ULTHO on the DMC benchmark.}
    \label{fig:sac_dmc}
\end{figure}

\clearpage\newpage

\section{Detailed Decision Processes}\label{appendix:decision}
\begin{figure*}[h!]
\centering
\includegraphics[width=\linewidth]{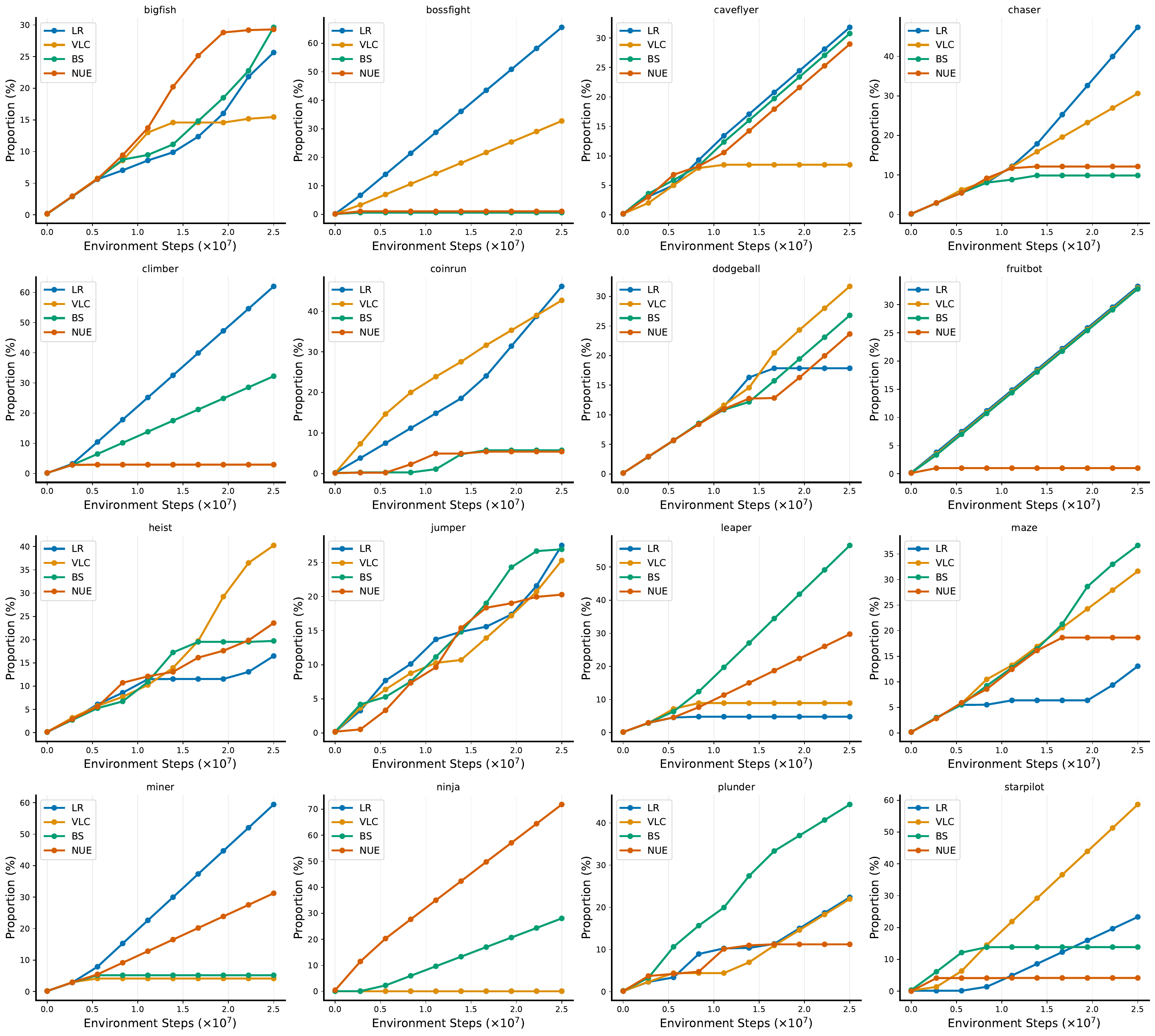}
\caption{Detailed decision processes of PPO+ULTHO on the Procgen benchmark.}
\label{fig:pg_ppo_inter_decisions_16}
\end{figure*}

\begin{figure*}[h!]
\centering
\includegraphics[width=\linewidth]{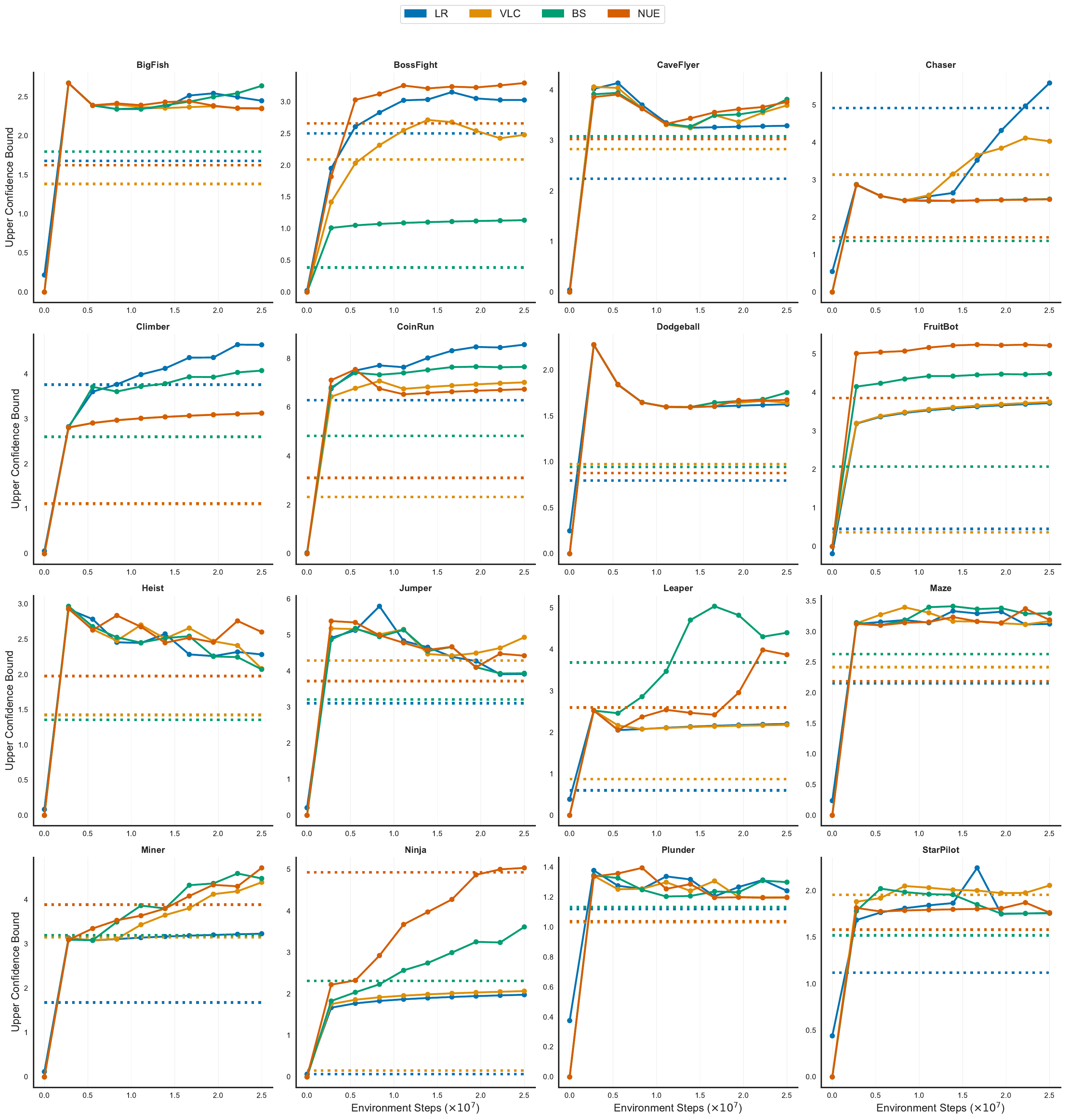}
\caption{The variation of confidence intervals of PPO+ULTHO on the Procgen benchmark. Here, the solid line represents the mean value, and the dashed line represents the final upper confidence bound.}
\label{fig:pg_ppo_inter_ct_16}
\end{figure*}



\clearpage\newpage

\end{document}